\documentclass{article}

\PassOptionsToPackage{square, numbers}{natbib}
\usepackage[square, numbers]{natbib}
 \usepackage[main, final]{neurips_2026}

\usepackage[utf8]{inputenc} 
\usepackage[T1]{fontenc}    
\usepackage{hyperref}       
\usepackage{url}            
\usepackage{makecell}
\usepackage{booktabs}       
\usepackage{amsfonts}       
\usepackage{nicefrac}       
\usepackage{microtype}      
\usepackage{xcolor}         
\usepackage{graphicx}
\usepackage{colortbl}
\usepackage{tcolorbox}
\usepackage{xcolor}
\newtheorem{observation}{Observation}
\usepackage{amsmath, amssymb}
\usepackage{multirow}
\usepackage{subcaption}
\title{Latent Abstraction for Retrieval-Augmented Generation}

\author{Ha Lan N.T\textsuperscript{*}, Minh-Anh Nguyen\textsuperscript{*}, Dung D. Le  \\
Center for AI Research, VinUniversity, Vietnam\\
\texttt{\{lan.nth, minh.na2, dung.ld\}@vinuni.edu.vn} \\
\textsuperscript{*}\textit{Equal contribution}
}

\begin{document}

\maketitle

\begin{abstract}
Retrieval-Augmented Generation (RAG) has become a standard approach for enhancing large language models (LLMs) with external knowledge, mitigating hallucinations, and improving factuality. However, existing systems rely on generating natural language queries at each hop and maintaining a strict architectural separation between retriever and generator, preventing them from leveraging the full representational capacity of the LLM. We propose \textbf{LAnR} (Latent Abstraction for RAG), a unified framework in which a single LLM jointly performs encoding, retrieval, and generation entirely within its own latent space. Rather than generating textual queries, LAnR produces dense retrieval vectors from the hidden states of a designated \texttt{[PRED]} token and uses them to match against encoded document representations from the same model. Furthermore, LAnR adaptively decides when sufficient evidence has been retrieved using a lightweight MLP control head over those same hidden states, eliminating explicit token-level stopping reasoning. Extensive experiments on five QA benchmarks spanning single-hop and multi-hop settings demonstrate that LAnR outperforms existing RAG methods, while achieving improved inference efficiency through reduced number of retrieval calls and tighter model integration.
\end{abstract}

\section{Introduction}

\begin{figure}[ht]
    \centering
    \includegraphics[width=\textwidth]{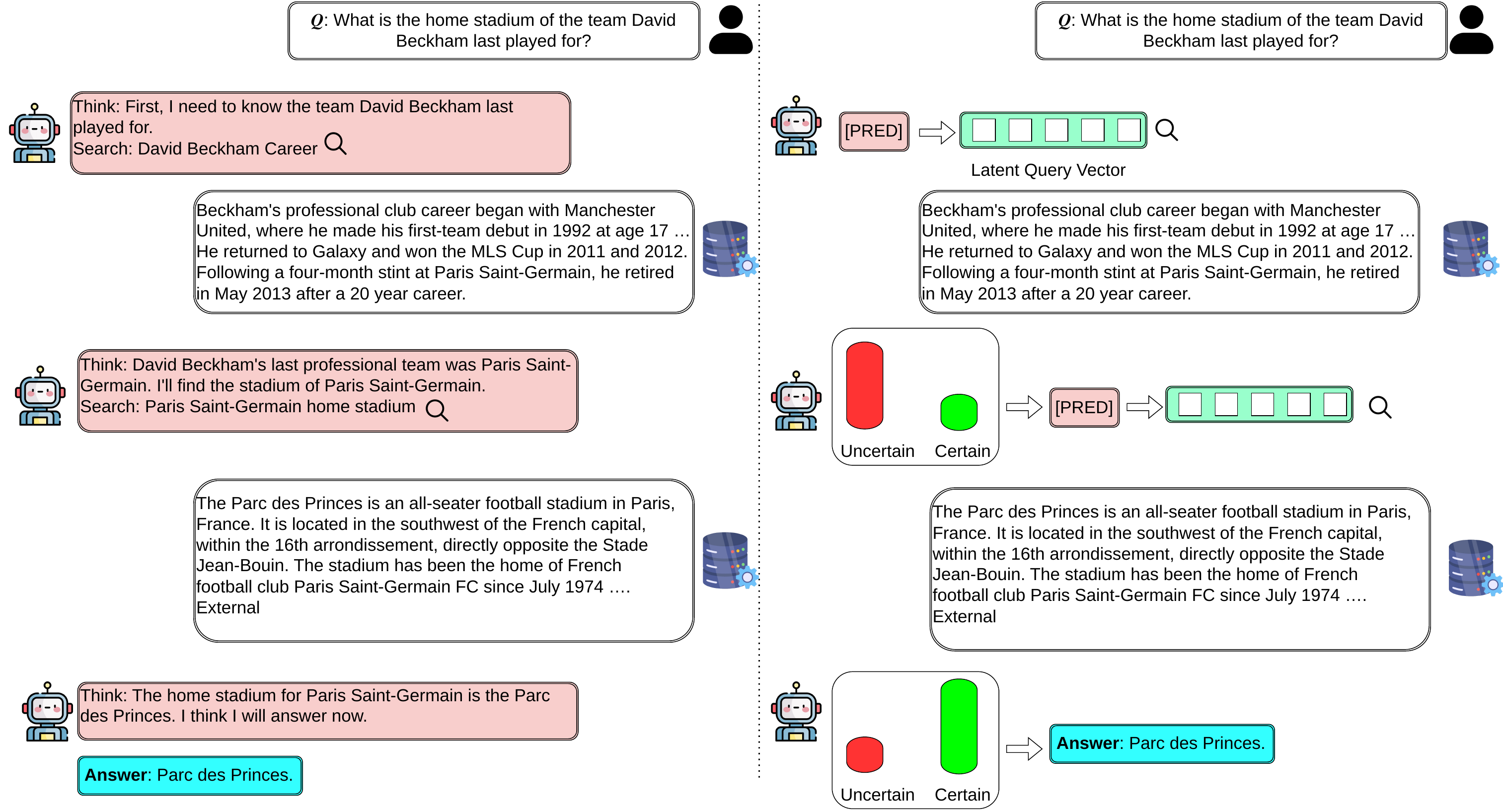}
\caption{\textbf{Comparison between conventional RAG and LAnR for multi-hop QA.} Conventional RAG performs explicit reasoning at each hop, including generating intermediate text, forming search queries, and deciding whether to continue retrieval. In contrast, LAnR operates in latent space: a special token \texttt{[PRED]} produces query vectors from hidden states, while a lightweight MLP controls the retrieval process, enabling more efficient and integrated reasoning.}
    \label{fig:comparison}
\end{figure}

Retrieval-Augmented Generation (RAG) has emerged as a standard paradigm for enhancing LLMs with external knowledge, improving factuality, and mitigating hallucinations \citep{asai2023retrieval, ge2023context, ram2023context}. By retrieving relevant documents from external corpora and conditioning generation on this information, RAG systems enable LLMs to access up-to-date and domain-specific knowledge beyond their parametric memory. Despite these advantages, existing RAG frameworks exhibit several limitations. First, they rely heavily on explicit text-based retrieval, where the model must generate natural language queries to interact with a separate retrieval module. Second, they enforce a strict architectural separation between the retriever and the generator, often requiring independently trained components or additional fine-tuning via reinforcement learning (RL) or supervised fine-tuning (SFT) LLMs to work with retrieval modules \citep{auto-refine,searchr1,ircot,self-rag}. These design choices introduce substantial computational overhead, limit the full utilization of the LLM’s shared representational capacity, and increase inference latency due to the need for explicit token-level generation at each step, illustrated in Figure~\ref{fig:comparison}.

\begin{figure}[ht]
    \centering
    \includegraphics[width=\textwidth]{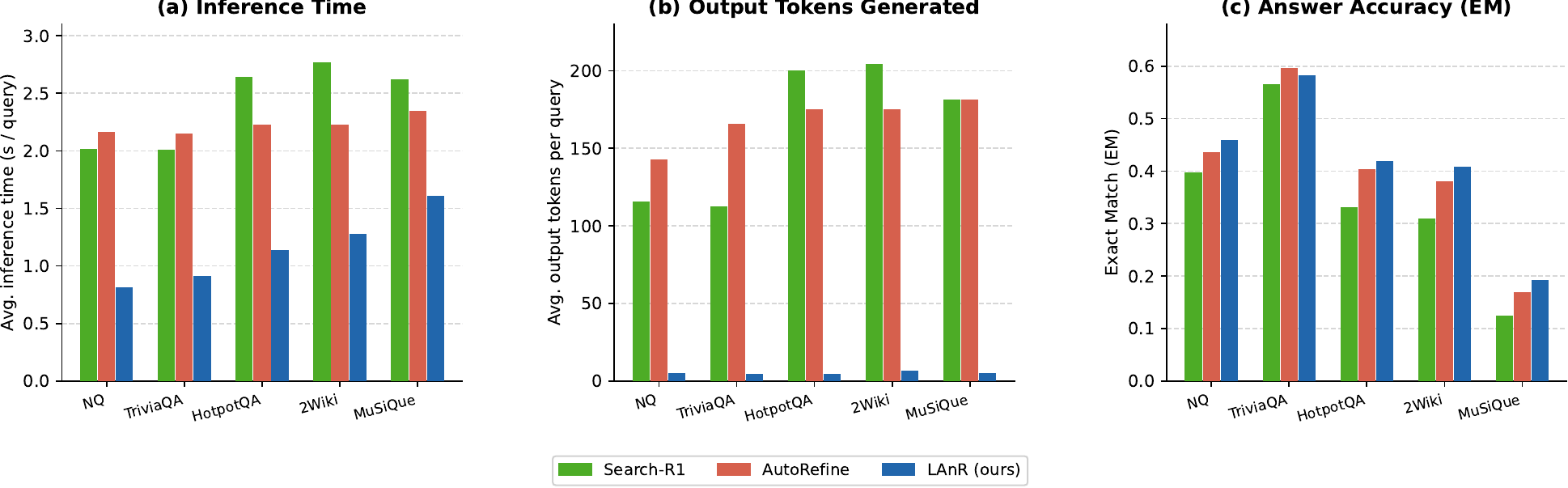}
\caption{Comparison of inference time, generated tokens, and Exact Match accuracy between prior RAG methods and LAnR. Latent retrieval reduces latency and token generation while maintaining strong performance. All methods use Qwen2.5-3B-Instruct as backbone model to ensure a fair comparison.}
    \label{fig:time_inference}
\end{figure}

Recent advances in latent reasoning suggest an alternative paradigm, where LLMs perform reasoning directly in hidden representation space rather than through fully verbalized intermediate steps \citep{hrpo,shi2025swireasoning,survey,xu2025softcot++}. Such approaches demonstrate that latent trajectories in hidden states can encode rich reasoning processes without explicit token generation. However, this paradigm remains largely unexplored in the context of retrieval-augmented systems \citep{survey}. Extending latent reasoning to RAG is, however, non-trivial. First, unlike standard reasoning tasks, RAG lacks high-quality chain-of-thought supervision for constructing retrieval queries, making it difficult to distill text-based reasoning into latent representations \citep{coconut,xu2025softcot}. Second, existing RAG architectures typically rely on separate embedding models for retrieval, which necessitate explicit natural language queries. Consequently, integrating both reasoning traces and retrieval queries into a shared latent space requires a unified model that simultaneously supports retrieval and generation.

To address these challenges, we propose a unified \textbf{latent abstraction retrieval-augmented generation} (LAnR) framework, wherein retrieval is conducted directly within the LLM’s internal representation space. Instead of generating textual queries, the model produces latent query vectors derived from the hidden states of designated special tokens \texttt{[PRED]}. These vectors are used to retrieve documents from a vector database constructed using representations from the same LLM, enabling tight integration between retrieval and generation, detailed in Section~\ref{sec:full_method}. To support multi-hop reasoning, we further introduce a lightweight multi-layer perceptron (MLP) \citep{noriega2005multilayer} head that predicts whether additional retrieval steps are necessary. This design is motivated by empirical observations that uncertainty signals, such as entropy in the token distribution, correlate with the need for further information acquisition, illustrated in Appendix ~\ref{sec:entropy}.

Extensive experiments across multiple benchmarks demonstrate that our approach achieves superior performance compared to state-of-the-art RAG methods, while significantly reducing inference latency due to minimal text generated, illustrated in Figure \ref{fig:time_inference} and eliminating the need for serving a separately trained dense retriever. We provide our code at the repository \url{https://github.com/akBear23/LAnR}. We summarize our main contributions as follows:
\begin{enumerate}
\item We propose a novel latent RAG framework that leverages the internal representation space of an LLM to jointly perform document encoding, retrieval, and generation, thereby unifying these components within a single architecture.
\item We design an implicit retrieval control mechanism, implemented as a lightweight auxiliary head, which uses the LLM's hidden representation to adaptively determine whether additional retrieval is needed, without relying on explicit text-based reasoning.
\item We introduce a new approach for constructing retrieval query vectors directly from the LLM's latent representations, eliminating the need for intermediate natural language query generation.
\end{enumerate}

\section{Latent Query Construction}
\label{sec:latent_query}

We begin by investigating latent query construction, where the LLM bypasses explicit textual query generation and instead produces a dense retrieval vector directly from its hidden representations. To enable this capability, the LLM is trained to internalize query formulation, allowing it to infer search intent from the input and map it directly into a retrieval vector. Throughout the paper, $\mathcal{M}$ is a Qwen2.5 backbone \citep{yang2025qwen3} (3B or 7B, Base or Instruct variant), which we extend with a special \texttt{[PRED]} token and a lightweight retrieval control head (Section~\ref{sec:full_method}); all parameters of the backbone are fine-tuned, and further implementation details are given in Appendix~\ref{appen:detail_training}. Formally, let $x = (x_1, \ldots, x_T)$ denote an input query token sequence and $\mathcal{M}$ an autoregressive language model with last-layer hidden states $h_t \in \mathbb{R}^d$. We append a designated token \texttt{[PRED]} to the input, forming $\tilde{x} = (x_1, \ldots, x_T, \texttt{[PRED]})$. Through causal self-attention, its hidden state attends to the full preceding context, yielding a latent query vector:
\begin{equation}
    q = h_{\texttt{[PRED]}} \in \mathbb{R}^d.
    \label{eq:latent_query}
\end{equation}
Optionally, $N \geq 1$ consecutive \texttt{[PRED]} tokens can be appended, allowing the model to construct the query representation over multiple latent steps, which promotes higher-level abstraction and extends the LLM’s latent reasoning process; we default to $N{=}1$ and study this design choice in Appendix~\ref{appen:more_results}.

Unlike prior LLM-based encoders \citep{llmembedding, llm2vec}, where the query vector encodes a fully-formed textual query, here $q$ is produced by the model's own reasoning over context. This distinction becomes essential in the multi-turn setting, detailed in Section~\ref{sec:full_method}, where intermediate sub-queries have no canonical textual form and must be inferred from the running context of previously retrieved evidence. Each representation of the document $D_i$ are obtained from the same model $\mathcal{M}$ via last-token pooling \citep{cheng2024xrag, ge2023context}:
\begin{equation}
    d_i = h_{D_i}^{\mathrm{last}} \in \mathbb{R}^d.
    \label{eq:doc_repr}
\end{equation}
Embedding queries and documents within a shared representation space induced by the same parameters eliminates the need for a separate retriever and preserves the model's generative capability for downstream answer production. We train the \texttt{[PRED]} representation with a standard contrastive objective \citep{E5, bge-m3}:
\begin{equation}
    \mathcal{L}_{\mathrm{CL}} = -\log \frac{\exp\bigl(\mathrm{sim}(q, d^+) / \tau\bigr)}{\exp\bigl(\mathrm{sim}(q, d^+) / \tau\bigr) + \sum_{j=1}^{N^-} \exp\bigl(\mathrm{sim}(q, d_j^-) / \tau\bigr)},
    \label{eq:contrastive}
\end{equation}
where $d^+$ is a positive document, $\{d_j^-\}_{j=1}^{N^-}$ are hard negatives mined from the model's own retrieval errors and periodically refreshed following ANCE \citep{xiong2020approximate}, $\mathrm{sim}(\cdot,\cdot)$ is cosine similarity, and $\tau$ is a temperature parameter. This setup gives us a query vector at any point in the model's forward pass. Building on this foundation, Section~\ref{sec:full_method} introduces a multi-turn training objective designed to leverage both properties effectively.

\begin{figure}[ht]
    \centering
    \includegraphics[width=\textwidth]{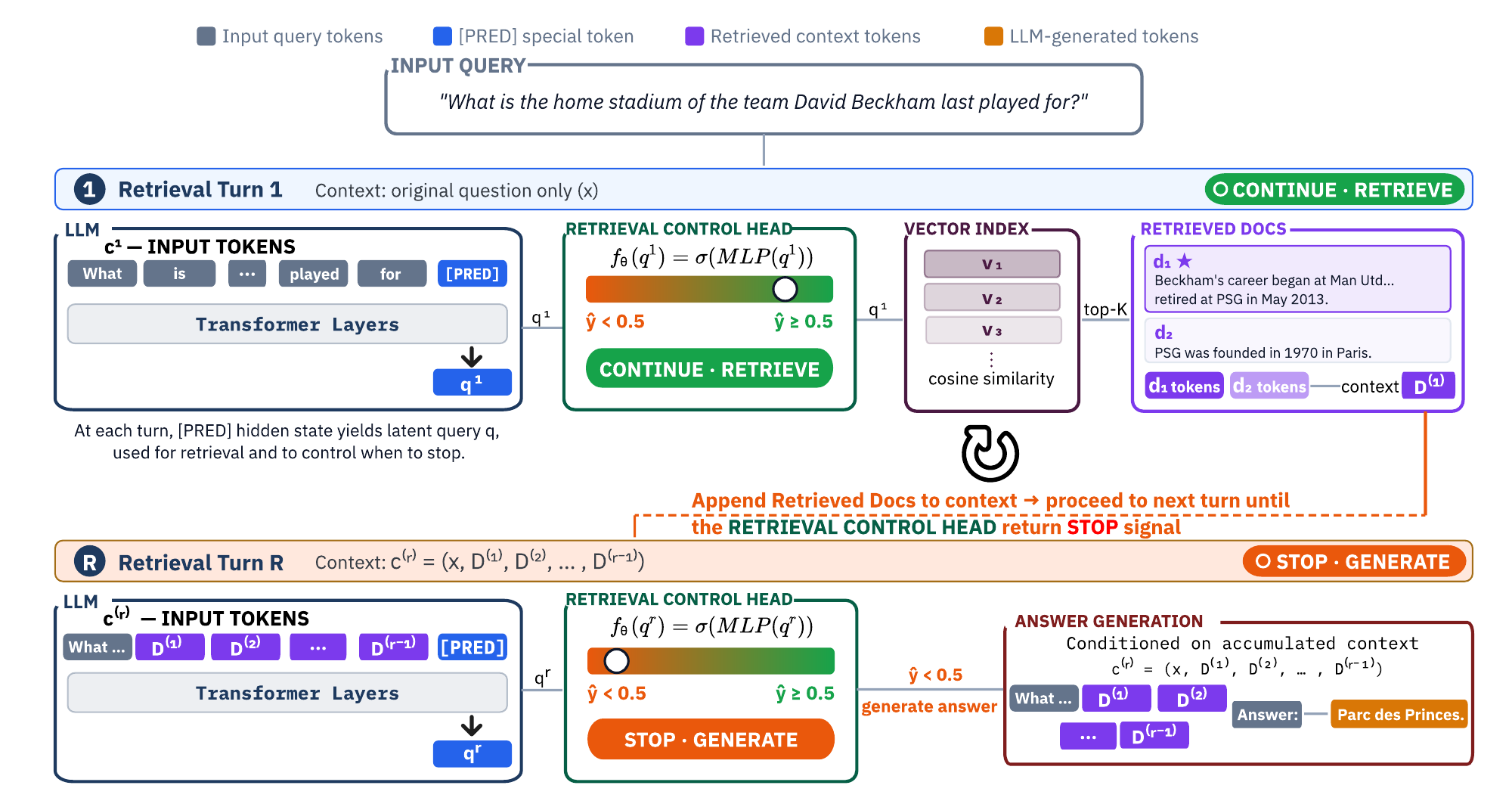}
    \caption{\textbf{Overview of LAnR.} Queries are injected into the LLM and combined with a \texttt{[PRED]} token to form a latent query from hidden representations. This latent query is used for retrieval and to decide whether further retrieval is needed via a lightweight MLP Retrieval Control Head. The LLM then generates the answer from the retrieved context.}
    \label{fig:overview}
\end{figure}

\section{End-to-End Framework and Training Objective}
\label{sec:full_method}

Building on the latent query construction in Section~\ref{sec:latent_query}, we extend the framework from single-step implicit retrieval to a multi-turn setting. The central challenge is to determine \emph{\textbf{when}} to trigger additional retrieval and \emph{\textbf{what}} to retrieve at each step, while relying solely on latent reasoning in hidden space without generating intermediate text tokens. We address both through: (1) an MLP-based retrieval control head that decides whether further retrieval is necessary, and (2) an adaptive contrastive target mechanism that dynamically updates the retrieval objective based on remaining unretrieved evidence.

At each retrieval turn $r = 1, 2, \ldots, R$, the model constructs a latent query from the current context, which now includes both the original question and any previously retrieved documents, and decides whether to retrieve again or to proceed with answer generation. Let $c^{(r)}$ denote the accumulated context at turn $r$:
\begin{equation*}
    c^{(r)} = \bigl(x,\; D^{(1)},\; D^{(2)},\; \ldots,\; D^{(r-1)}\bigr),
    \label{eq:context}
\end{equation*}
where $D^{(s)}$ denotes the set of top-$K$ documents retrieved at turn $s$. At each turn, we append \texttt{[PRED]} to $c^{(r)}$ and extract the latent query vector: $q^{(r)} = h_{\texttt{[PRED]}}^{(r)} \in \mathbb{R}^d$, which is used both for retrieval via similarity matching against the document index, detailed in Equation~\ref{eq:doc_repr} and as input to the retrieval control head described below. The process repeats until the control head signals termination or a maximum number of turns $R$ is reached, after which the model generates the final answer conditioned on the full accumulated context $c^{(R+1)}$.

\textbf{MLP-based Retrieval Control Head.}\quad An additional component of our framework is a lightweight MLP head that determines whether the currently retrieved evidence is sufficient to answer the query, or whether additional retrieval is required. At each turn $r$, the Retrieval Control Head $f_\theta$ takes $q^{(r)}$ as input and produces a binary decision:
\begin{equation}
    \hat{y}^{(r)} = f_\theta\bigl(q^{(r)}\bigr) = \sigma\bigl(\mathrm{MLP}(q^{(r)})\bigr) \in [0, 1],
    \label{eq:control_head}
\end{equation}
where $\sigma$ is the sigmoid function. A prediction $\hat{y}^{(r)} < 0.5$ indicates that all necessary evidence has been retrieved and the model should proceed to generation; $\hat{y}^{(r)} \geq 0.5$ signals that additional retrieval is needed. The training label is derived from the model's own retrieval state. Let $\mathcal{P}$ denote the full set of positive (gold) documents required to answer the query, and $\mathcal{R}^{(r)} = D^{(1)} \cup \cdots \cup D^{(r)}$ the documents retrieved up to turn $r$. The ground-truth label is:
\begin{equation}
    y^{(r)} = 
    \begin{cases}
        0, & \text{if } \mathcal{P} \subseteq \mathcal{R}^{(r)}, \\
        1, & \text{otherwise},
    \end{cases}
    \label{eq:control_label}
\end{equation}
i.e., the label is $0$ (stop) when all positive documents have been successfully retrieved, and $1$ (continue) otherwise. This label requires no additional stopping supervision: it is computed automatically from the model's own retrieval trajectory by checking whether the retrieved set covers $\mathcal{P}$, reusing the same evidence annotations already required to train the retriever (Eq.~\ref{eq:adaptive_cl}). No manually annotated stopping decisions, retrieval trajectories, or external LLM judges are needed. For datasets without annotated supporting passages, $\mathcal{P}$ is constructed by the same distant-supervision procedure used for retriever training (Appendix~\ref{appen:control_supervision}). The control head is trained with binary cross-entropy:
\begin{equation}
    \mathcal{L}_{\mathrm{ctrl}} = -\sum_{r=1}^{R} \Bigl[ y^{(r)} \log \hat{y}^{(r)} + (1 - y^{(r)}) \log (1 - \hat{y}^{(r)}) \Bigr].
    \label{eq:control_loss}
\end{equation}

In the multi-turn setting, naively using the same contrastive target at every turn is suboptimal: once a positive document has been retrieved, it should no longer serve as the target for subsequent queries. At turn $r$, the set of leftover positive documents is: $\mathcal{P}^{(r)} = \mathcal{P} \setminus \mathcal{R}^{(r-1)}$,
where $\mathcal{R}^{(0)} = \emptyset$. The contrastive loss at turn $r$ uses a positive document sampled from $\mathcal{P}^{(r)}$:
\begin{equation}
    \mathcal{L}_{\mathrm{CL}}^{(r)} = -\log \frac{\exp\bigl(\mathrm{sim}(q^{(r)}, d^{+(r)}) / \tau\bigr)}{\exp\bigl(\mathrm{sim}(q^{(r)}, d^{+(r)}) / \tau\bigr) + \sum_{j=1}^{N^-} \exp\bigl(\mathrm{sim}(q^{(r)}, d_j^-) / \tau\bigr)},
    \label{eq:adaptive_cl}
\end{equation}
where $d^{+(r)}$ is the embedding of a document drawn from $\mathcal{P}^{(r)}$. This steers each successive turn toward the missing evidence, avoiding redundant retrieval. The full training loss combines next-token prediction, multi-turn contrastive retrieval, and the control head objective:
\begin{equation}
    \mathcal{L} = \sum_{r=1}^{R} \mathcal{L}_{\mathrm{CL}}^{(r)} + \lambda \,\mathcal{L}_{\mathrm{NTP}} + \mu\, \mathcal{L}_{\mathrm{ctrl}},
    \label{eq:full_loss}
\end{equation}
where $\lambda$ and $\mu$ are hyperparameters balancing the three objectives. $\mathcal{L}_{\mathrm{NTP}}$ denotes the standard next-token prediction loss, which maintains the generative capability of the LLM. We apply loss masking to retrieved tokens and \texttt{[PRED]} tokens, so that the optimization objective is computed only over tokens generated by the LLM, excluding retrieved content from gradient updates, following prior works \citep{searchr1,auto-refine}.


\textbf{Inference.} At inference time, the contrastive and control losses are not used; the trained model directly produces latent queries and stopping decisions. Retrieval starts from the input question $x$. At each turn $r$, the model appends a \texttt{[PRED]} token to the current context $c^{(r)}$ and runs a single forward pass to obtain $q^{(r)}=h_{\texttt{[PRED]}}^{(r)}$. The retrieval control head then computes $\hat{y}^{(r)}=f_\theta(q^{(r)})$. If $\hat{y}^{(r)} \geq 0.5$, $q^{(r)}$ is used as the query for nearest-neighbour search against the document index built with Eq.~\ref{eq:doc_repr}; the top-$K$ documents are appended to the context, and the next turn begins. If $\hat{y}^{(r)} < 0.5$, retrieval terminates. Importantly, no text is generated during the retrieval phase: the model produces neither intermediate reasoning tokens nor textual search queries, only latent \texttt{[PRED]} states. After retrieval terminates (or the maximum of $R=4$ rounds is reached), the prompt \texttt{Answer:} is appended to the accumulated context, and the model decodes the final answer autoregressively in a single generation pass. Figure~\ref{fig:overview} illustrates this procedure, and Appendix~\ref{appen:case_study} provides hop-by-hop qualitative examples.

\section{Experiments}\label{sec:experiment}
In this section, we investigate the following research questions (RQs):

\textbf{RQ1.} How effective is LAnR compared to conventional agentic RAG systems in answer quality across single-hop and multi-hop benchmarks, and can its implicit retriever match text embedding models in retrieval accuracy?

\textbf{RQ2.} Does LAnR's retrieval control head adaptively allocate search hops based on query complexity, and does each retrieval hop find relevant evidence more effectively than text-based iterative search?

\textbf{RQ3.} Can LAnR reduce inference time by minimizing retrieval iterations and improving efficiency through more abstract query representations?


\textbf{RQ4.} What is the individual contribution of each training objective ($\mathcal{L}_\text{NTP}$, $\mathcal{L}_\text{CL}$, $\mathcal{L}_\text{ctrl}$) and the retrieval control head to LAnR's end-to-end performance?

\subsection{Experimental Setup}

\paragraph{Datasets.} We evaluate our method on five open-domain QA benchmarks, including two single-hop datasets, Natural Questions (NQ) \cite{kwiatkowski2019natural} and TriviaQA \cite{joshi2017triviaqa}, and three multi-hop datasets: HotpotQA \cite{yang2018hotpotqa}, 2WikiMultihopQA (2Wiki) \cite{ho2020constructing}, and MuSiQue \cite{trivedi2022musique}. We report Exact Match (EM) as the primary metric for answer generation. For retrieval evaluation on datasets with ground-truth documents, we use Recall as the main metric.

\paragraph{Baselines.} We compare LAnR against three categories of methods: (1) generation without retrieval, including direct LLM generation and supervised fine-tuning (SFT) without retrieval; (2) single-hop retrieval methods, such as naive RAG \cite{lewis2020retrieval} that retrieves once using the input query; and (3) multi-hop retrieval approaches, including agentic and iterative systems such as Search-o1 \cite{searcho1}, IRCoT \cite{ircot}, Search-R1 \cite{searchr1}, and AutoRefine \cite{auto-refine}.

\paragraph{Implementation Details.} To simulate a realistic retrieval setting, we remove ground-truth context from the QA datasets and instead use the 2018 Wikipedia dump \cite{karpukhin2020dense} as the external knowledge source. By default, retrieval returns the top-$k$ documents at each step, with $k=3$ following the setting of Search-R1 for fair comparison. All models are trained on the combined NQ and HotpotQA dataset, following the setup of prior works \cite{searchr1, auto-refine}. Unless otherwise specified, all baselines and LAnR use the same Qwen2.5-3B models to ensure a fair comparison. Detailed training configurations are described in Appendix~\ref{appen:detail_training}.

\begin{table*}[t]
\centering
\setlength{\tabcolsep}{4pt}   
\renewcommand{\arraystretch}{0.85}  
\caption{\textbf{(RQ1)} Accuracy comparison of LAnR and baselines on QA benchmarks. \textbf{Bold} and \underline{underline} denote best and second-best results, respectively. "-Instruct" and "-Base" indicate the corresponding Qwen2.5-3B backbone variants.}
\label{tab:main_results}
\small
\begin{tabular}{lcccccc}
\toprule
& \multicolumn{2}{c}{Single-Hop QA} & \multicolumn{3}{c}{Multi-Hop QA} & \\
\cmidrule(lr){2-3} \cmidrule(lr){4-6}
Methods & NQ & TriviaQA & HotpotQA & 2Wiki & Musique & Avg. \\
\midrule
\multicolumn{7}{l}{\textit{w/o Retrieval}} \\
\quad Direct Generation & 0.106 & 0.288 & 0.149 & 0.244 & 0.020 & 0.161 \\
\quad SFT               & 0.249 & 0.292 & 0.186 & 0.248 & 0.044 & 0.204 \\
\midrule
\multicolumn{7}{l}{\textit{w/ Single-Hop Retrieval}} \\
\quad Naive RAG~\cite{lewis2020retrieval} & 0.348 & 0.544 & 0.255 & 0.226 & 0.047 & 0.284 \\
\midrule
\multicolumn{7}{l}{\textit{w/ Multi-Hop Retrieval}} \\
\quad Search-o1~\cite{searcho1}         & 0.238 & 0.472 & 0.221 & 0.218 & 0.054 & 0.241 \\
\quad IRCoT~\cite{ircot}               & 0.111 & 0.312 & 0.164 & 0.171 & 0.067 & 0.165 \\
\quad ReSearch-Instruct~\cite{research} & 0.365 & 0.571 & 0.351 & 0.272 & 0.095 & 0.331 \\
\quad ReSearch-Base~\cite{research}     & 0.427 & 0.597 & 0.305 & 0.272 & 0.074 & 0.335 \\
\quad Search-R1-Instruct~\cite{searchr1} & 0.397 & 0.565 & 0.331 & 0.310 & 0.124 & 0.345 \\
\quad Search-R1-Base~\cite{searchr1}    & 0.421 & 0.583 & 0.297 & 0.274 & 0.066 & 0.328 \\
\quad AutoRefine-Instruct \cite{auto-refine} & 0.436 & 0.597 & 0.404 & 0.380 & 0.169 & 0.397 \\
\quad AutoRefine-Base \cite{auto-refine}    & \textbf{0.467} & \textbf{0.620} & 0.405 & 0.393 & 0.157 & 0.408 \\
\midrule
\rowcolor{gray!15}
\quad LAnR-Instruct     & \underline{0.460} & \underline{0.613} & \textbf{0.419} & \textbf{0.408} & \textbf{0.193} & \textbf{0.419} \\
\rowcolor{gray!15}
\quad LAnR-Base         & 0.455 & 0.610 & \underline{0.417} & \underline{0.402} & \underline{0.187} & \underline{0.414} \\
\bottomrule
\end{tabular}%
\end{table*}

\subsection{Main Results}

\paragraph{Overall Performance (RQ1).} Table~\ref{tab:main_results} reports EM accuracy across five QA benchmarks covering both single-hop and multi-hop reasoning. Retrieval-free methods perform poorly, while Naive RAG improves single-hop QA but remains ineffective on compositional tasks such as HotpotQA, 2Wiki, and MuSiQue. Among multi-hop approaches, Search-R1 and AutoRefine provide strong baselines, but LAnR consistently achieves the best overall performance, especially on challenging multi-hop benchmarks. LAnR-Instruct reaches 0.419 EM on HotpotQA, 0.408 on 2Wiki, and 0.193 on MuSiQue, outperforming AutoRefine-Instruct across all three datasets. Although AutoRefine remains competitive on single-hop benchmarks, LAnR achieves the highest average EM overall, indicating a stronger balance between retrieval and reasoning. Despite being trained only on NQ and HotpotQA, LAnR also generalizes effectively to unseen datasets in a zero-shot setting. 

\begin{table}[ht]
\centering
\small
\setlength{\tabcolsep}{4pt}
\caption{%
  \textbf{Retrieval recall at varying document budgets.}
  BM25 is an unsupervised lexical retriever; BGE and E5 are dense retrievers trained with retrieval supervision.
  LAnR-Instruct-static uses the single-step \texttt{[PRED]} embedding as a fixed query vector (Section~\ref{sec:latent_query});
  LAnR-Instruct uses adaptive multi-hop retrieval with the learned controller, and the last column reports its recall together with the average number of retrieved documents.
  Despite joint training with generation, LAnR-Instruct exceeds the R@5 of all retrievers using a comparable number of documents. Matched-budget comparisons are given in Table~\ref{tab:matched_budget}.
}
\label{tab:retrieval_recall}
\begin{tabular}{@{}llccccc@{}}
\toprule
Dataset & Method & R@1 & R@3 & R@5 & R@10 & Recall (budget) \\
\midrule
\multirow{5}{*}{HotpotQA}
 & BM25~\cite{robertson2009probabilistic} & 0.401 & 0.603 & 0.670 & 0.751 & --- \\
 & BGE~\cite{bge-m3}                      & 0.477 & 0.782 & 0.832 & 0.880 & --- \\
 & E5~\cite{E5}                           & 0.462 & 0.773 & 0.826 & 0.875 & --- \\
 & LAnR-Instruct-static                          & 0.451 & 0.769 & 0.800 & 0.830 & --- \\
 & LAnR-Instruct  & ---   & ---   & ---   & ---   & 0.840{\scriptsize$\pm$0.0012} (\textasciitilde5.5 docs) \\
\midrule
\multirow{5}{*}{2WikiMQA}
 & BM25~\cite{robertson2009probabilistic} & 0.360 & 0.555 & 0.605 & 0.656 & --- \\
 & BGE~\cite{bge-m3}                      & 0.411 & 0.643 & 0.680 & 0.715 & --- \\
 & E5~\cite{E5}                           & 0.415 & 0.656 & 0.687 & 0.717 & --- \\
 & LAnR-Instruct-static                          & 0.407 & 0.641 & 0.684 & 0.713 & --- \\
 & LAnR-Instruct & ---   & ---   & ---   & ---   & 0.715{\scriptsize$\pm$0.0009} (\textasciitilde5.1 docs) \\
\midrule
\multirow{5}{*}{MuSiQue}
 & BM25~\cite{robertson2009probabilistic} & 0.216 & 0.310 & 0.350 & 0.400 & --- \\
 & BGE~\cite{bge-m3}                      & 0.280 & 0.435 & 0.500 & 0.573 & --- \\
 & E5~\cite{E5}                           & 0.290 & 0.417 & 0.454 & 0.503 & --- \\
 & LAnR-Instruct-static   & 0.265 & 0.393 & 0.449 & 0.487 & --- \\
 & LAnR-Instruct   & ---   & ---   & ---   & ---   & 0.516{\scriptsize$\pm$0.0011} (\textasciitilde7.9 docs) \\
\bottomrule
\end{tabular}
\end{table}

\begin{figure}[ht]
    \centering
    \includegraphics[width=\textwidth]{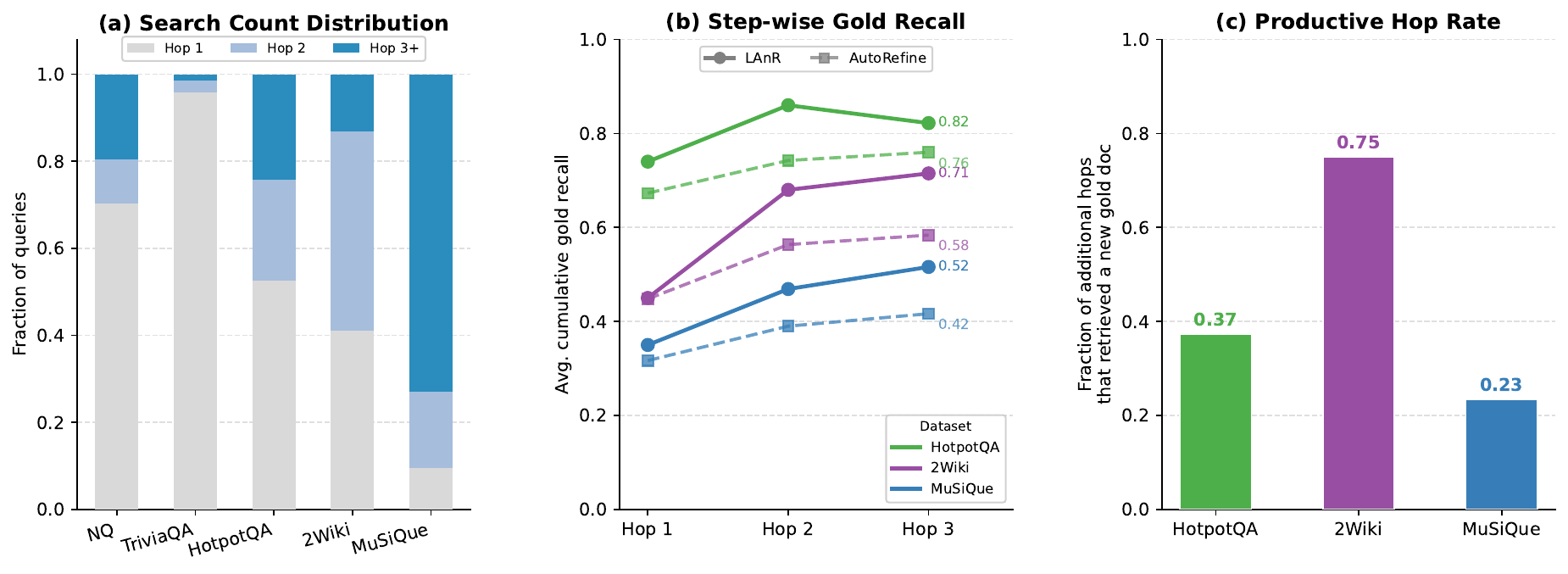}
    \caption{
        \textbf{(RQ2) LAnR adaptive search behaviour.}
        \textbf{(a)} Distribution of search counts per query across all five benchmarks.
        Single-hop datasets are dominated by a single search call,
        while compositional benchmarks require progressively more,
        showing the control head adapts to query complexity rather than applying a fixed budget.
        \textbf{(b)} Step-wise cumulative gold recall at each hop for LAnR (solid) and
        AutoRefine (dashed) on HotpotQA, 2Wiki, and MuSiQue.
        LAnR consistently outperforms AutoRefine at every hop on all three datasets.
        \textbf{(c)} Productive hop rate: fraction of additional search calls ($\geq$ Hop~2)
        that retrieved at least one new gold document per dataset.
    }
    \label{fig:search_behavior}
\end{figure}

\paragraph{Retrieval Quality vs. Text Embedding Retrievers.} Table~\ref{tab:retrieval_recall} compares LAnR-Instruct with sparse and dense retrievers across three multi-hop QA benchmarks. While BM25, BGE, and E5 rely on fixed top-$K$ retrieval and are trained solely for retrieval, LAnR-Instruct jointly performs adaptive retrieval and generation. Despite this joint objective, LAnR-Instruct achieves competitive retrieval recall using substantially fewer retrieved documents. On HotpotQA and 2WikiMQA, it attains recalls of 0.840 and 0.715 with only $\sim$5 retrieved documents, matching or exceeding the Recall@5--10 performance of specialized retrievers. On the more challenging MuSiQue benchmark, LAnR-Instruct achieves 0.516 recall with $\sim$7.9 documents, outperforming BM25 and remaining competitive with dense retrievers. In contrast, the single-step LAnR-Instruct-static variant consistently underperforms the adaptive version, highlighting the benefit of iterative retrieval with learned control.

\paragraph{Latent RAG Effectiveness (RQ2).} To understand \emph{how} LAnR retrieves rather than just \emph{how well}, we analyse the search-level behaviour of the retrieval control head across all five evaluation benchmarks.

\textit{Adaptive search count.} Figure~\ref{fig:search_behavior}(a) shows the distribution of search calls per query across all five datasets. On single-hop benchmarks (NQ, TriviaQA), the vast majority of queries are resolved in a single search call, reflecting the control head's ability to recognise when sufficient evidence has been gathered early. On compositional multi-hop benchmarks the distribution shifts substantially: 2Wiki and MuSiQue queries spread across two and three search calls, with MuSiQue requiring the most, consistent with its longer evidence chains.

\textit{Step-wise retrieval quality.} Figure~\ref{fig:search_behavior}(b) compares cumulative gold recall across retrieval hops between LAnR and AutoRefine on the three multi-hop benchmarks. LAnR consistently achieves higher recall at every step, with the advantage appearing from the first hop and widening on more compositional datasets such as 2Wiki and MuSiQue. This suggests that latent query vectors better capture residual information needs, enabling subsequent retrieval steps to target missing evidence more effectively than text-based queries.

\textit{Control head precision.}
Figure~\ref{fig:search_behavior}(c) reports the productive hop rate: the fraction of additional search calls triggered by a "continue" decision that successfully retrieved at least one new gold document. On 2Wiki, 75\% of additional calls are productive, confirming that the control head issues follow-up searches only when evidence is genuinely incomplete. On MuSiQue, the rate is lower at 23\%, reflecting the difficulty of locating the precise gold documents in highly compositional three-hop and four-hop chains where missing any single query may fail to bridge the full reasoning gap.

\begin{table}[ht]
\centering
\small
\setlength{\tabcolsep}{5pt}
\caption{%
  \textbf{Retrieval recall under matched document budgets.}
  \textbf{(a)} 6-document budget: BM25, BGE, E5, and LAnR-static each retrieve the top-6 documents in a single call, while LAnR-adaptive performs two latent retrieval calls of top-3 documents each.
  \textbf{(b)} 10-document budget: retrievers use top-10, while LAnR-adaptive retrieves exactly 10 unique documents through three latent calls (top-3, top-3, top-4).
  The total number of retrieved documents is identical within each panel; the number of retrieval calls differs.
  \textbf{Bold} marks the best result per row.
}
\label{tab:matched_budget}
\begin{tabular}{@{}lccccc@{}}
\toprule
Dataset & BM25 & BGE & E5 & LAnR-static & LAnR-adaptive \\
\midrule
\multicolumn{6}{@{}l}{\textit{(a) 6-document budget} \quad (LAnR-adaptive: $2\times$top-3)} \\
\addlinespace[2pt]
NQ        & 0.638 & 0.798 & \textbf{0.818} & 0.812 & 0.795 \\
TriviaQA  & 0.641 & 0.751 & \textbf{0.811} & 0.805 & 0.798 \\
HotpotQA  & 0.692 & 0.847 & 0.841 & 0.847 & \textbf{0.865} \\
2Wiki     & 0.620 & 0.690 & 0.696 & 0.693 & \textbf{0.724} \\
MuSiQue   & 0.362 & 0.509 & 0.468 & 0.460 & \textbf{0.520} \\
\midrule
\multicolumn{6}{@{}l}{\textit{(b) 10-document budget} \quad (LAnR-adaptive: top-3 + top-3 + top-4)} \\
\addlinespace[2pt]
HotpotQA  & 0.751 & 0.880 & 0.875 & 0.830 & \textbf{0.905} \\
2Wiki     & 0.656 & 0.715 & 0.717 & 0.713 & \textbf{0.828} \\
MuSiQue   & 0.400 & 0.573 & 0.503 & 0.487 & \textbf{0.601} \\
\bottomrule
\end{tabular}
\end{table}
 
\paragraph{Matched-budget retrieval (RQ2).} Table~\ref{tab:matched_budget} controls for retrieval volume by fixing the total number of documents. With a budget of six documents, LAnR-adaptive ($2\times$top-3) outperforms LAnR-static (top-6) and conventional retrievers on all three multi-hop benchmarks, achieving 0.520 on MuSiQue versus 0.460 and 0.509 for LAnR-static and BGE, respectively. With ten documents, it reaches 0.828 on 2Wiki, surpassing E5 (0.717). These results highlight the benefit of iterative latent queries in targeting residual evidence. On single-hop tasks, a forced second retrieval slightly reduces recall, as it may introduce irrelevant documents after sufficient evidence has been found. The default controller mitigates this by typically stopping after one retrieval call (Figure~\ref{fig:search_behavior}a). Note that document budgets, but not retrieval-call counts, are matched; latency is evaluated separately in Figure~\ref{fig:time_inference}.

\begin{figure}[ht]
    \centering
    \includegraphics[width=\textwidth]{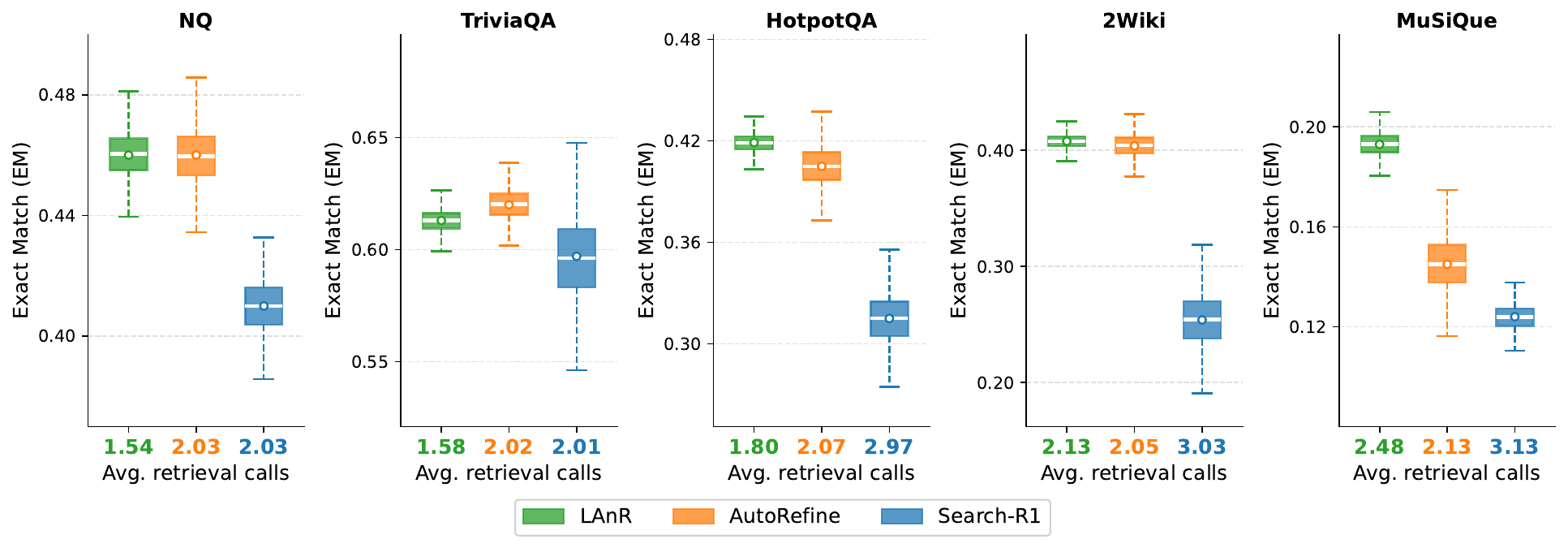}
    \caption{\textbf{(RQ3)} Per-dataset EM distributions for LAnR, AutoRefine, and Search-R1. LAnR achieves competitive or higher EM with the fewest retrieval calls and consistently narrow variance.}
    \label{fig:search_box}
\end{figure}

\paragraph{Inference Efficiency of Latent Retrieval (RQ3).}
Figure~\ref{fig:search_box} directly tests the core efficiency claim of LAnR: each subplot shows the EM distribution for all three methods on a single benchmark, with the colored x-tick labels reporting each method's average number of retrieval calls. Across all five datasets, LAnR issues the fewest retrieval calls ($1.54-2.47$), yet its median EM matches or exceeds AutoRefine ($2.02-2.13$ calls) and Search-R1 ($2.01-3.13$ calls). The narrow interquartile ranges confirm that LAnR's accuracy is stable across evaluation instances rather than driven by a subset of easy queries. Notably, on the multihop benchmarks, LAnR's gap over Search-R1 widens substantially despite Search-R1 issuing nearly twice as many retrievals, indicating that latent query vectors retrieve more relevant evidence per call than token-based queries.

The token and latency costs are quantified in Figure~\ref{fig:time_inference}.
LAnR generates on average only $\sim$5 output tokens per query, roughly $30\times$ fewer than Search-R1 ($\sim$163) or AutoRefine ($\sim$168), because the retrieval signal is encoded as a latent vector rather than verbalized text. This reduction translates directly to wall-clock savings: LAnR completes a query in $0.81-1.61$\,s compared to $2.01-2.77$\,s for Search-R1 and $2.15-2.34\,$s for AutoRefine, a $1.5-2.7\times$ speedup across all benchmarks. 

\begin{tcolorbox}[
    colback=blue!3!white,
    colframe=blue!35!black,
    boxrule=0.6pt,
    arc=2mm,
    left=1mm,
    right=1mm,
    top=1mm,
    bottom=1mm
]
\textbf{Key Takeaway.} By replacing explicit query generation with compact latent vectors and an adaptive retrieval controller, LAnR performs fewer but more effective search calls, achieves higher cumulative gold recall, and maintains stable EM performance with lower computational cost.
\end{tcolorbox}

\subsection{Ablation Studies}

\paragraph{Component ablation (\textbf{RQ4}).} Table~\ref{tab:component_ablation} isolates the contribution of three design choices: the MLP retrieval control head, NTP loss masking, and the NTP loss weight $\lambda_{\mathrm{NTP}}$. Removing the control head and replacing it with a fixed 5-document budget leaves single-hop performance essentially unchanged but causes large drops on multi-hop tasks, confirming that adaptive retrieval is the primary driver of multi-hop gains. Removing NTP loss masking introduces a moderate but consistent degradation on all three multi-hop benchmarks, suggesting the masking prevents the model from conflating retrieval-query generation with next-token prediction over retrieved content. Reducing $\lambda_{\mathrm{NTP}}$ to $0.5$ incurs a small performance drop, while setting it to $0$ collapses performance to near zero, demonstrating that the NTP objective is indispensable for answer generation.

\begin{table}[ht]
\centering
\caption{Component ablation study (\textbf{RQ4}). $\Delta$Avg denotes the absolute decrease in average performance relative to the full LAnR model.}
\label{tab:component_ablation}
\small
\setlength{\tabcolsep}{5pt}
\begin{tabular}{lccccccr}
\toprule
Variant & NQ & TriviaQA & HotpotQA & 2Wiki & MuSiQue & Avg & $\Delta$Avg \\
\midrule
\rowcolor{gray!15}
\textbf{LAnR-Instruct} & \textbf{0.460} & \textbf{0.613} & \textbf{0.419} & \textbf{0.408} & \textbf{0.193} & \textbf{0.419} & --- \\
\midrule
w/o ctrl head & 0.459 & 0.592 & 0.342 & 0.266 & 0.095 & 0.351 & $-0.068$ \\
w/o NTP loss masking        & 0.458 & 0.582 & 0.395 & 0.371 & 0.182 & 0.398 & $-0.021$ \\
$\lambda_{\mathrm{NTP}} = 0.5$ & 0.455 & 0.578 & 0.411 & 0.402 & 0.189 & 0.407 & $-0.012$ \\
$\lambda_{\mathrm{NTP}} = 0$ & 0.005 & 0.007 & 0.004 & 0.000 & 0.000 & 0.003 & $-0.416$ \\
\bottomrule
\end{tabular}
\end{table}

\begin{figure}[ht]
     \centering
    \includegraphics[width=\textwidth]{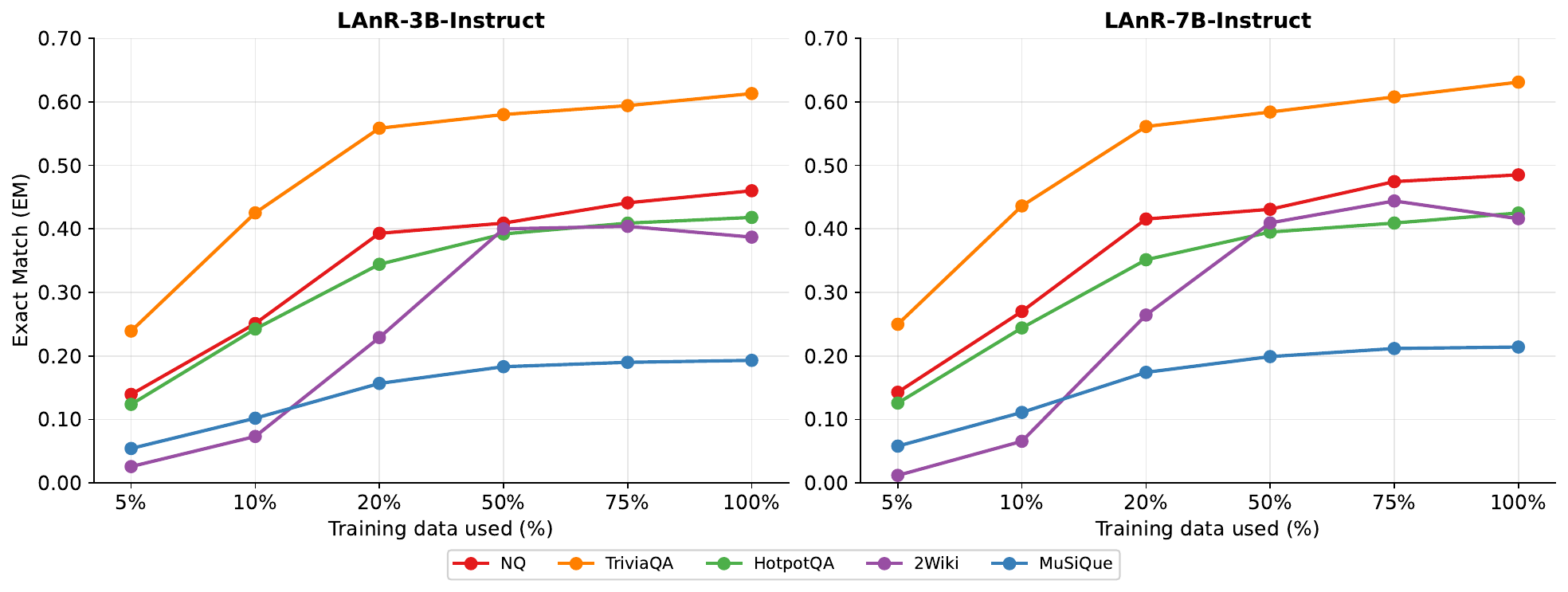}
    \caption{EM at various fractions of training data used (5\,\% to 100\,\%). Most of the performance gain is captured by the first 50\,\% of the data.}
    \label{fig:data_size_ablation}
\end{figure}

\paragraph{Additional Experimental Results.} Figure~\ref{fig:data_size_ablation} shows EM as a function of training-set fraction for both LAnR-3B and LAnR-7B across all five benchmarks.
The learning curve rises steeply in the first 10-20\% of data, where the four-dataset average EM climbs from 0.13 to 0.38.
Notably, 2WikiMultihopQA starts near zero at 5\% and gains most rapidly between 10-50\%, reflecting the higher sample complexity of multi-hop compositional reasoning compared to single-hop benchmarks (NQ, TriviaQA), which plateau earlier.
Beyond 50\% gains flatten to within 3-5\% of the full-data ceiling, suggesting that LAnR's latent retrieval mechanism generalises effectively with moderate training budgets.
The 7B variant yields a consistent improvement over 3B across all data fractions, indicating that model scale and training data contribute largely independently to final performance.

Additional analyses are provided to further examine LAnR from multiple perspectives, including: (i) a detailed evaluation of the Retrieval Control Head, Appendix~\ref{appen:retrieval_control_head_analysis}; (ii) model scaling experiments spanning 3B to 7B backbones, Appendix~\ref{appen:model_size}; (iii) statistical significance analysis across multiple random seeds, Appendix~\ref{appen:variance_analysis}; and (iv) the effect of varying the number of \texttt{[PRED]} tokens, where larger token counts improve abstraction capacity, Appendix~\ref{appen:PRED_numers}. Together, these findings further demonstrate the robustness, scalability, and retrieval effectiveness of LAnR across diverse QA benchmarks.

\section{Conclusion}

We introduced \textbf{latent abstraction retrieval-augmented generation} (LAnR), a unified framework that performs retrieval and reasoning directly in the latent space of a single LLM. By replacing explicit text-based query generation with latent query vectors derived from hidden states, LAnR eliminates the need for a separate retriever and reduces reliance on token-level reasoning. We further proposed a lightweight retrieval control head that adaptively determines when additional retrieval is required, enabling efficient multi-hop reasoning without explicit intermediate text. Empirical results across multiple benchmarks demonstrate that LAnR achieves competitive performance compared to existing RAG systems while significantly reducing inference latency. These findings highlight the potential of latent-space reasoning as a scalable alternative to conventional RAG pipelines.

\newpage
\bibliographystyle{plainnat}
\bibliography{ref}

@article{cheng2024xrag,
  title={xrag: Extreme context compression for retrieval-augmented generation with one token},
  author={Cheng, Xin and Wang, Xun and Zhang, Xingxing and Ge, Tao and Chen, Si-Qing and Wei, Furu and Zhang, Huishuai and Zhao, Dongyan},
  journal={Advances in Neural Information Processing Systems},
  volume={37},
  pages={109487--109516},
  year={2024}
}

@article{ge2023context,
  title={In-context autoencoder for context compression in a large language model},
  author={Ge, Tao and Hu, Jing and Wang, Lei and Wang, Xun and Chen, Si-Qing and Wei, Furu},
  journal={arXiv preprint arXiv:2307.06945},
  year={2023}
}

@article{searchr1,
  title={Search-r1: Training llms to reason and leverage search engines with reinforcement learning},
  author={Jin, Bowen and Zeng, Hansi and Yue, Zhenrui and Yoon, Jinsung and Arik, Sercan and Wang, Dong and Zamani, Hamed and Han, Jiawei},
  journal={arXiv preprint arXiv:2503.09516},
  year={2025}
}

@article{auto-refine,
  title={Search and Refine During Think: Facilitating Knowledge Refinement for Improved Retrieval-Augmented Reasoning},
  author={Shi, Yaorui and Li, Sihang and Wu, Chang and Liu, Zhiyuan and Fang, Junfeng and Cai, Hengxing and Zhang, An and Wang, Xiang},
  journal={arXiv preprint arXiv:2505.11277},
  year={2025}
}

@article{xiong2020approximate,
  title={Approximate nearest neighbor negative contrastive learning for dense text retrieval},
  author={Xiong, Lee and Xiong, Chenyan and Li, Ye and Tang, Kwok-Fung and Liu, Jialin and Bennett, Paul and Ahmed, Junaid and Overwijk, Arnold},
  journal={arXiv preprint arXiv:2007.00808},
  year={2020}
}

@inproceedings{asai2023retrieval,
  title={Retrieval-based language models and applications},
  author={Asai, Akari and Min, Sewon and Zhong, Zexuan and Chen, Danqi},
  booktitle={Proceedings of the 61st Annual Meeting of the Association for Computational Linguistics (Volume 6: Tutorial Abstracts)},
  pages={41--46},
  year={2023}
}

@article{ram2023context,
  title={In-context retrieval-augmented language models},
  author={Ram, Ori and Levine, Yoav and Dalmedigos, Itay and Muhlgay, Dor and Shashua, Amnon and Leyton-Brown, Kevin and Shoham, Yoav},
  journal={Transactions of the Association for Computational Linguistics},
  volume={11},
  pages={1316--1331},
  year={2023},
  publisher={MIT Press One Broadway, 12th Floor, Cambridge, Massachusetts 02142, USA~…}
}

@inproceedings{self-rag,
  title={Self-rag: Learning to retrieve, generate, and critique through self-reflection},
  author={Asai, Akari and Wu, Zeqiu and Wang, Yizhong and Sil, Avirup and Hajishirzi, Hannaneh},
  booktitle={The Twelfth International Conference on Learning Representations},
  year={2023}
}

@inproceedings{ircot,
  title={Interleaving retrieval with chain-of-thought reasoning for knowledge-intensive multi-step questions},
  author={Trivedi, Harsh and Balasubramanian, Niranjan and Khot, Tushar and Sabharwal, Ashish},
  booktitle={Proceedings of the 61st annual meeting of the association for computational linguistics (volume 1: long papers)},
  pages={10014--10037},
  year={2023}
}

@article{coconut,
  title={Training large language models to reason in a continuous latent space},
  author={Hao, Shibo and Sukhbaatar, Sainbayar and Su, DiJia and Li, Xian and Hu, Zhiting and Weston, Jason and Tian, Yuandong},
  journal={arXiv preprint arXiv:2412.06769},
  year={2024}
}

@article{hrpo,
  title={Hybrid latent reasoning via reinforcement learning},
  author={Yue, Zhenrui and Jin, Bowen and Zeng, Huimin and Zhuang, Honglei and Qin, Zhen and Yoon, Jinsung and Shang, Lanyu and Han, Jiawei and Wang, Dong},
  journal={arXiv preprint arXiv:2505.18454},
  year={2025}
}

@article{shi2025swireasoning,
  title={SwiReasoning: Switch-Thinking in Latent and Explicit for Pareto-Superior Reasoning LLMs},
  author={Shi, Dachuan and Asi, Abedelkadir and Li, Keying and Yuan, Xiangchi and Pan, Leyan and Lee, Wenke and Xiao, Wen},
  journal={arXiv preprint arXiv:2510.05069},
  year={2025}
}

@inproceedings{xu2025softcot,
  title={Softcot: Soft chain-of-thought for efficient reasoning with llms},
  author={Xu, Yige and Guo, Xu and Zeng, Zhiwei and Miao, Chunyan},
  booktitle={Proceedings of the 63rd Annual Meeting of the Association for Computational Linguistics (Volume 1: Long Papers)},
  pages={23336--23351},
  year={2025}
}

@article{noriega2005multilayer,
  title={Multilayer perceptron tutorial},
  author={Noriega, Leonardo},
  journal={School of Computing. Staffordshire University},
  volume={4},
  number={5},
  pages={444},
  year={2005}
}

@article{E5,
  title={Text embeddings by weakly-supervised contrastive pre-training},
  author={Wang, Liang and Yang, Nan and Huang, Xiaolong and Jiao, Binxing and Yang, Linjun and Jiang, Daxin and Majumder, Rangan and Wei, Furu},
  journal={arXiv preprint arXiv:2212.03533},
  year={2022}
}

@article{bge-m3,
  title={Bge m3-embedding: Multi-lingual, multi-functionality, multi-granularity text embeddings through self-knowledge distillation},
  author={Chen, Jianlv and Xiao, Shitao and Zhang, Peitian and Luo, Kun and Lian, Defu and Liu, Zheng},
  journal={arXiv preprint arXiv:2402.03216},
  volume={4},
  number={5},
  year={2024}
}

@article{wei2022chain,
  title={Chain-of-thought prompting elicits reasoning in large language models},
  author={Wei, Jason and Wang, Xuezhi and Schuurmans, Dale and Bosma, Maarten and Xia, Fei and Chi, Ed and Le, Quoc V and Zhou, Denny and others},
  journal={Advances in neural information processing systems},
  volume={35},
  pages={24824--24837},
  year={2022}
}

@article{yao2023tree,
  title={Tree of thoughts: Deliberate problem solving with large language models},
  author={Yao, Shunyu and Yu, Dian and Zhao, Jeffrey and Shafran, Izhak and Griffiths, Tom and Cao, Yuan and Narasimhan, Karthik},
  journal={Advances in neural information processing systems},
  volume={36},
  pages={11809--11822},
  year={2023}
}

@article{goyal2023think,
  title={Think before you speak: Training language models with pause tokens},
  author={Goyal, Sachin and Ji, Ziwei and Rawat, Ankit Singh and Menon, Aditya Krishna and Kumar, Sanjiv and Nagarajan, Vaishnavh},
  journal={arXiv preprint arXiv:2310.02226},
  year={2023}
}

@article{wang2025system,
  title={System-1.5 reasoning: Traversal in language and latent spaces with dynamic shortcuts},
  author={Wang, Xiaoqiang and Wang, Suyuchen and Zhu, Yun and Liu, Bang},
  journal={arXiv preprint arXiv:2505.18962},
  year={2025}
}

@article{chen2025reasoning,
  title={Reasoning beyond language: A comprehensive survey on latent chain-of-thought reasoning},
  author={Chen, Xinghao and Zhao, Anhao and Xia, Heming and Lu, Xuan and Wang, Hanlin and Chen, Yanjun and Zhang, Wei and Wang, Jian and Li, Wenjie and Shen, Xiaoyu},
  journal={arXiv preprint arXiv:2505.16782},
  year={2025}
}

@article{li2025implicit,
  title={Implicit reasoning in large language models: A comprehensive survey},
  author={Li, Jindong and Fu, Yali and Fan, Li and Liu, Jiahong and Shu, Yao and Qin, Chengwei and Yang, Menglin and King, Irwin and Ying, Rex},
  journal={arXiv preprint arXiv:2509.02350},
  year={2025}
}

@article{lewis2020retrieval,
  title={Retrieval-augmented generation for knowledge-intensive nlp tasks},
  author={Lewis, Patrick and Perez, Ethan and Piktus, Aleksandra and Petroni, Fabio and Karpukhin, Vladimir and Goyal, Naman and K{\"u}ttler, Heinrich and Lewis, Mike and Yih, Wen-tau and Rockt{\"a}schel, Tim and others},
  journal={Advances in neural information processing systems},
  volume={33},
  pages={9459--9474},
  year={2020}
}

@article{yue2024inference,
  title={Inference scaling for long-context retrieval augmented generation},
  author={Yue, Zhenrui and Zhuang, Honglei and Bai, Aijun and Hui, Kai and Jagerman, Rolf and Zeng, Hansi and Qin, Zhen and Wang, Dong and Wang, Xuanhui and Bendersky, Michael},
  journal={arXiv preprint arXiv:2410.04343},
  year={2024}
}

@article{arslan2024survey,
  title={A Survey on RAG with LLMs},
  author={Arslan, Muhammad and Ghanem, Hussam and Munawar, Saba and Cruz, Christophe},
  journal={Procedia computer science},
  volume={246},
  pages={3781--3790},
  year={2024},
  publisher={Elsevier}
}

@inproceedings{jiang2023active,
  title={Active retrieval augmented generation},
  author={Jiang, Zhengbao and Xu, Frank F and Gao, Luyu and Sun, Zhiqing and Liu, Qian and Dwivedi-Yu, Jane and Yang, Yiming and Callan, Jamie and Neubig, Graham},
  booktitle={Proceedings of the 2023 conference on empirical methods in natural language processing},
  pages={7969--7992},
  year={2023}
}

@article{jin2024long,
  title={Long-context llms meet rag: Overcoming challenges for long inputs in rag},
  author={Jin, Bowen and Yoon, Jinsung and Han, Jiawei and Arik, Sercan O},
  journal={arXiv preprint arXiv:2410.05983},
  year={2024}
}

@article{chen2025learning,
  title={Learning to reason with search for llms via reinforcement learning},
  author={Chen, Mingyang and Sun, Linzhuang and Li, Tianpeng and Sun, Haoze and Zhou, Yijie and Zhu, Chenzheng and Wang, Haofen and Pan, Jeff Z and Zhang, Wen and Chen, Huajun and others},
  journal={arXiv preprint arXiv:2503.19470},
  year={2025}
}

@inproceedings{yao2022react,
  title={React: Synergizing reasoning and acting in language models},
  author={Yao, Shunyu and Zhao, Jeffrey and Yu, Dian and Du, Nan and Shafran, Izhak and Narasimhan, Karthik R and Cao, Yuan},
  booktitle={The eleventh international conference on learning representations},
  year={2022}
}

@article{schick2023toolformer,
  title={Toolformer: Language models can teach themselves to use tools},
  author={Schick, Timo and Dwivedi-Yu, Jane and Dess{\`\i}, Roberto and Raileanu, Roberta and Lomeli, Maria and Hambro, Eric and Zettlemoyer, Luke and Cancedda, Nicola and Scialom, Thomas},
  journal={Advances in neural information processing systems},
  volume={36},
  pages={68539--68551},
  year={2023}
}

@inproceedings{searcho1,
  title={Search-o1: Agentic search-enhanced large reasoning models},
  author={Li, Xiaoxi and Dong, Guanting and Jin, Jiajie and Zhang, Yuyao and Zhou, Yujia and Zhu, Yutao and Zhang, Peitian and Dou, Zhicheng},
  booktitle={Proceedings of the 2025 Conference on Empirical Methods in Natural Language Processing},
  pages={5420--5438},
  year={2025}
}

@article{research,
  title={Learning to reason with search for llms via reinforcement learning},
  author={Chen, Mingyang and Sun, Linzhuang and Li, Tianpeng and Sun, Haoze and Zhou, Yijie and Zhu, Chenzheng and Wang, Haofen and Pan, Jeff Z and Zhang, Wen and Chen, Huajun and others},
  journal={arXiv preprint arXiv:2503.19470},
  year={2025}
}

@article{yang2025qwen3,
  title={Qwen3 technical report},
  author={Yang, An and Li, Anfeng and Yang, Baosong and Zhang, Beichen and Hui, Binyuan and Zheng, Bo and Yu, Bowen and Gao, Chang and Huang, Chengen and Lv, Chenxu and others},
  journal={arXiv preprint arXiv:2505.09388},
  year={2025}
}

@inproceedings{karpukhin2020dense,
  title={Dense passage retrieval for open-domain question answering},
  author={Karpukhin, Vladimir and Oguz, Barlas and Min, Sewon and Lewis, Patrick and Wu, Ledell and Edunov, Sergey and Chen, Danqi and Yih, Wen-tau},
  booktitle={Proceedings of the 2020 conference on empirical methods in natural language processing (EMNLP)},
  pages={6769--6781},
  year={2020}
}

@article{survey,
  title={Reasoning beyond language: A comprehensive survey on latent chain-of-thought reasoning},
  author={Chen, Xinghao and Zhao, Anhao and Xia, Heming and Lu, Xuan and Wang, Hanlin and Chen, Yanjun and Zhang, Wei and Wang, Jian and Li, Wenjie and Shen, Xiaoyu},
  journal={arXiv preprint arXiv:2505.16782},
  year={2025}
}

@article{xu2025softcot++,
  title={Softcot++: Test-time scaling with soft chain-of-thought reasoning},
  author={Xu, Yige and Guo, Xu and Zeng, Zhiwei and Miao, Chunyan},
  journal={arXiv preprint arXiv:2505.11484},
  year={2025}
}

@inproceedings{llmembedding,
  title={Improving text embeddings with large language models},
  author={Wang, Liang and Yang, Nan and Huang, Xiaolong and Yang, Linjun and Majumder, Rangan and Wei, Furu},
  booktitle={Proceedings of the 62nd Annual Meeting of the Association for Computational Linguistics (Volume 1: Long Papers)},
  pages={11897--11916},
  year={2024}
}

@article{llm2vec,
  title={Llm2vec: Large language models are secretly powerful text encoders, 2024},
  author={BehnamGhader, Parishad and Adlakha, Vaibhav and Mosbach, Marius and Bahdanau, Dzmitry and Chapados, Nicolas and Reddy, Siva},
  journal={URL https://arxiv. org/abs/2404.05961},
  year={2024}
}

@article{yu2026latent,
  title={The latent space: Foundation, evolution, mechanism, ability, and outlook},
  author={Yu, Xinlei and Chen, Zhangquan and He, Yongbo and Fu, Tianyu and Yang, Cheng and Xu, Chengming and Ma, Yue and Hu, Xiaobin and Cao, Zhe and Xu, Jie and others},
  journal={arXiv preprint arXiv:2604.02029},
  year={2026}
}

@inproceedings{das2025rader,
  title={Rader: Reasoning-aware dense retrieval models},
  author={Das, Debrup and O’Nuallain, Sam and Rahimi, Razieh},
  booktitle={Proceedings of the 2025 Conference on Empirical Methods in Natural Language Processing},
  pages={19981--20008},
  year={2025}
}

@article{springer2024repetition,
  title={Repetition improves language model embeddings},
  author={Springer, Jacob Mitchell and Kotha, Suhas and Fried, Daniel and Neubig, Graham and Raghunathan, Aditi},
  journal={arXiv preprint arXiv:2402.15449},
  year={2024}
}

@inproceedings{deng2025following,
  title={Following the autoregressive nature of llm embeddings via compression and alignment},
  author={Deng, Jingcheng and Jiang, Zhongtao and Pang, Liang and Wei, Zihao and Chen, Liwei and Xu, Kun and Song, Yang and Shen, Huawei and Cheng, Xueqi},
  booktitle={Proceedings of the 2025 Conference on Empirical Methods in Natural Language Processing},
  pages={12672--12688},
  year={2025}
}

@article{nie2024text,
  title={When text embedding meets large language model: a comprehensive survey},
  author={Nie, Zhijie and Feng, Zhangchi and Li, Mingxin and Zhang, Cunwang and Zhang, Yanzhao and Long, Dingkun and Zhang, Richong},
  journal={arXiv preprint arXiv:2412.09165},
  year={2024}
}

@article{silva2024improving,
  title={Improving dense retrieval models with LLM augmented data for dataset search},
  author={Silva, Levy and Barbosa, Luciano},
  journal={Knowledge-based systems},
  volume={294},
  pages={111740},
  year={2024},
  publisher={Elsevier}
}

@article{kwiatkowski2019natural,
  title={Natural questions: a benchmark for question answering research},
  author={Kwiatkowski, Tom and Palomaki, Jennimaria and Redfield, Olivia and Collins, Michael and Parikh, Ankur and Alberti, Chris and Epstein, Danielle and Polosukhin, Illia and Devlin, Jacob and Lee, Kenton and others},
  journal={Transactions of the Association for Computational Linguistics},
  volume={7},
  pages={453--466},
  year={2019},
  publisher={MIT Press One Rogers Street, Cambridge, MA 02142-1209, USA journals-info~…}
}

@inproceedings{joshi2017triviaqa,
  title={Triviaqa: A large scale distantly supervised challenge dataset for reading comprehension},
  author={Joshi, Mandar and Choi, Eunsol and Weld, Daniel S and Zettlemoyer, Luke},
  booktitle={Proceedings of the 55th Annual Meeting of the Association for Computational Linguistics (Volume 1: Long Papers)},
  pages={1601--1611},
  year={2017}
}

@inproceedings{yang2018hotpotqa,
  title={HotpotQA: A dataset for diverse, explainable multi-hop question answering},
  author={Yang, Zhilin and Qi, Peng and Zhang, Saizheng and Bengio, Yoshua and Cohen, William and Salakhutdinov, Ruslan and Manning, Christopher D},
  booktitle={Proceedings of the 2018 conference on empirical methods in natural language processing},
  pages={2369--2380},
  year={2018}
}

@inproceedings{ho2020constructing,
  title={Constructing a multi-hop qa dataset for comprehensive evaluation of reasoning steps},
  author={Ho, Xanh and Nguyen, Anh-Khoa Duong and Sugawara, Saku and Aizawa, Akiko},
  booktitle={Proceedings of the 28th International Conference on Computational Linguistics},
  pages={6609--6625},
  year={2020}
}

@article{trivedi2022musique,
  title={MuSiQue: Multihop Questions via Single-hop Question Composition},
  author={Trivedi, Harsh and Balasubramanian, Niranjan and Khot, Tushar and Sabharwal, Ashish},
  journal={Transactions of the Association for Computational Linguistics},
  volume={10},
  pages={539--554},
  year={2022},
  publisher={MIT Press One Broadway, 12th Floor, Cambridge, Massachusetts 02142, USA~…}
}

@book{robertson2009probabilistic,
  title={The probabilistic relevance framework: BM25 and beyond},
  author={Robertson, Stephen and Zaragoza, Hugo},
  volume={4},
  year={2009},
  publisher={Now Publishers Inc}
}

@article{kim2025freeson,
  title={FREESON: Retriever-Free Retrieval-Augmented Reasoning via Corpus-Traversing MCTS},
  author={Kim, Chaeeun and Kim, Seungone},
  journal={arXiv preprint arXiv:2505.16409},
  year={2025}
}

@inproceedings{huanshuo2025ctrla,
  title={Ctrla: Adaptive retrieval-augmented generation via inherent control},
  author={Huanshuo, Liu and Zhang, Hao and Guo, Zhijiang and Wang, Jing and Dong, Kuicai and Li, Xiangyang and Lee, Yi Quan and Zhang, Cong and Liu, Yong},
  booktitle={Findings of the Association for Computational Linguistics: ACL 2025},
  pages={12592--12618},
  year={2025}
}

\appendix

\section{Related Works}

\subsection{Latent Reasoning}

A key limitation of dominant reasoning paradigms such as explicit chain-of-thought (CoT) \cite{wei2022chain,yao2023tree,goyal2023think,yu2026latent} lies in their reliance on discrete token generation during inference. In standard CoT decoding, the model commits to a single token at each step, sampled from the predicted distribution. While this process enhances interpretability by verbalizing intermediate reasoning steps, it inherently collapses the full probability distribution into a single trajectory, discarding uncertainty and eliminating alternative reasoning paths that may be informative. To address this limitation, recent work has explored latent reasoning \cite{coconut,xu2025softcot,shi2025swireasoning,hrpo,wang2025system}, where reasoning unfolds directly in the continuous hidden-state space rather than through discrete text. This paradigm offers two primary advantages over CoT: (1) increased representational capacity per step, as continuous vectors can encode substantially richer information than individual tokens \cite{chen2025reasoning}; and (2) the ability to implicitly preserve multiple reasoning hypotheses without prematurely collapsing them into a single token sequence \cite{li2025implicit}. However, despite these advances, prior work on latent reasoning has primarily focused on improving reasoning quality in standalone language modeling settings. Its application to RAG remains largely unexplored. In particular, existing methods do not address how latent representations can be leveraged for retrieval itself, nor how they can guide adaptive retrieval decisions. 

\subsection{Retrieval Augmented Generation}
RAG enhances LLMs by incorporating external knowledge sources to mitigate hallucinations and address knowledge gaps \cite{lewis2020retrieval,yue2024inference,arslan2024survey}. A central challenge in RAG systems lies in determining when and how to retrieve relevant information, as naive single-step retrieval often introduces irrelevant or insufficient context \cite{jiang2023active,jin2024long}. Early approaches employ supervised fine-tuning (SFT) to train models for query generation and retrieval integration \cite{self-rag,ircot,yao2022react,schick2023toolformer}; however, these methods rely on high-quality annotated trajectories and often exhibit limited generalization to out-of-distribution settings. Leveraging the inherent structure of LLMs, an alternative direction is to fine-tune them as document encoders for text embedding \cite{llm2vec,llmembedding,springer2024repetition,deng2025following}, or to use LLMs to generate synthetic data to support embedding training \cite{das2025rader,nie2024text,silva2024improving}. More recent work explores iterative and adaptive retrieval strategies, where models interleave reasoning and retrieval in a multi-step process, commonly described as "search-during-think" \cite{searchr1,auto-refine,chen2025learning,kim2025freeson}. However, existing approaches primarily operate in the discrete text space, relying on explicit query generation and token-level signals to control retrieval. Moreover, they often lack mechanisms for directly assessing retrieval sufficiency or refining retrieved evidence beyond surface-level interactions. 

\section{More Implementation Details}\label{appen:detail_training}

\begin{table}[ht]
\centering
\caption{Primary hyperparameters used by LAnR.}
\label{tab:hyperparams}
\begin{tabular}{lc}
\toprule
\textbf{Hyper-parameter} & \textbf{Value} \\
\midrule
Training Batch Size             &  256 \\
Chunk size                      & 100 words \\
Micro Training Batch Size       &  32\\
Learning Rate                   & $5 \times 10^{-5}$ \\
Validation Batch Size           & 256 \\
Total Training Steps            & 250 \\
Max Search Actions              &  4 \\
$\lambda$ for NTP Loss          & 1   \\
$\mu$ for Retrieval Control Head Loss & 1   \\
\bottomrule
\end{tabular}
\end{table}

\begin{table}[ht]
\centering
\caption{Statistics of the datasets used in this paper.}
\label{tab:dataset_stats}
\begin{tabular}{lrrrrr}
\toprule
       & NQ    & TriviaQA & HotpotQA & 2Wiki  & Musique \\
\midrule
Train  & 79168 & 78785    & 90447    & 15,000 & 19,938  \\
Dev    & 8757  & 8837     & 7405     & 12576  & 2417    \\
Test   & 3610  & 11313    & -        & -      & -       \\
\bottomrule
\end{tabular}
\end{table}

\begin{table}[ht]
\centering
\small
\setlength{\tabcolsep}{6pt}
\caption{%
\textbf{Serving efficiency comparison.}
Both Auto-Refine-Instruct and LAnR-Instruct use the same Qwen2.5-3B backbone.
Auto-Refine-Instruct requires an additional external embedding model for retrieval, whereas LAnR-Instruct performs retrieval and generation within a unified model.
}
\label{tab:serving_efficiency}
\begin{tabular}{lccc}
\toprule
Method & Retrieval Architecture & Extra Retriever & VRAM Usage (GB BF16)\\
\midrule
Auto-Refine-Instruct & Text-based retrieval & Yes & 6.48 \\
LAnR-Instruct & Unified latent retrieval & No & 5.85 \\
\bottomrule
\end{tabular}
\end{table}

\paragraph{Training Details} We train LAnR using full-parameter fine-tuning on 2 NVIDIA H100 (80GB) GPUs. The training data is constructed by combining Natural Questions (NQ) \cite{kwiatkowski2019natural} and HotpotQA \cite{yang2018hotpotqa}, following a consistent setup across LAnR and all training-based baselines. We employ Fully Sharded Data Parallelism (FSDP) for distributed training and use bfloat16 precision for both training and evaluation. Table~\ref{tab:hyperparams} summarizes the main hyperparameters. During inference, we sample with a temperature of 1.0 and allow up to 4 retrieval steps per query. Retrieved documents are concatenated and truncated to a maximum length of 512 tokens. For direct inference and supervised fine-tuning (SFT) baselines, we adopt Qwen2.5-3B-Instruct \cite{yang2025qwen3} as the backbone model.

\paragraph{Dataset Statistics.} All datasets are obtained from the FlashRAG Datasets collection\footnote{\url{https://huggingface.co/datasets/RUC-NLPIR/FlashRAG_datasets}}. Detailed statistics are reported in Table~\ref{tab:dataset_stats}. The LAnR training set is constructed from the training splits of NQ and HotpotQA, comprising 169,615 examples. For evaluation, we aggregate the test or development splits from seven benchmarks: test splits are used for datasets that provide them (e.g., NQ, TriviaQA), while development splits are used otherwise (e.g., HotpotQA, 2Wiki, MuSiQue).



\paragraph{Computational serving.}
Although LAnR-Instruct introduces higher indexing cost than conventional text embedding approaches, it reduces serving-time computational overhead by unifying retrieval and generation within a single model. Table~\ref{tab:serving_efficiency} compares LAnR-Instruct with Auto-Refine-Instruct using the same Qwen2.5-3B backbone. Unlike Auto-Refine-Instruct, which requires an additional external embedding retriever during inference, LAnR-Instruct performs retrieval directly through latent representations inside the generation model. As a result, LAnR-Instruct requires lower VRAM usage and simplifies deployment by eliminating the need to maintain a separate text retrieval encoder.

\subsection{Retrieval Control Head Model Design}

The retrieval control head $f_\theta$ is a single affine projection followed by a sigmoid:
\begin{equation}
    \hat{y}^{(r)} = \sigma\!\bigl(W\, q^{(r)} + b\bigr), \qquad
    W \in \mathbb{R}^{1 \times d},\quad b \in \mathbb{R},
    \label{eq:ctrl_arch}
\end{equation}
where $q^{(r)} = h_{\texttt{[PRED]}}^{(r)} \in \mathbb{R}^{d}$ is the last-layer hidden state of the final \texttt{[PRED]} token at retrieval turn $r$, and $d$ is the LLM's hidden dimension ($d{=}2048$ for Qwen-2.5-3B, $d{=}3584$ for Qwen-2.5-7B). The head introduces $d{+}1$ trainable parameters, 2{,}049 for the 3B backbone and 3{,}585 for the 7B backbone, adding negligible capacity (under $0.0002\%$ of total model parameters) and requiring zero additional LLM forward passes at inference.

\section{Latent Retrieval Control} \label{sec:entropy}

\begin{figure}[ht]
    \centering
    \includegraphics[width=\textwidth]{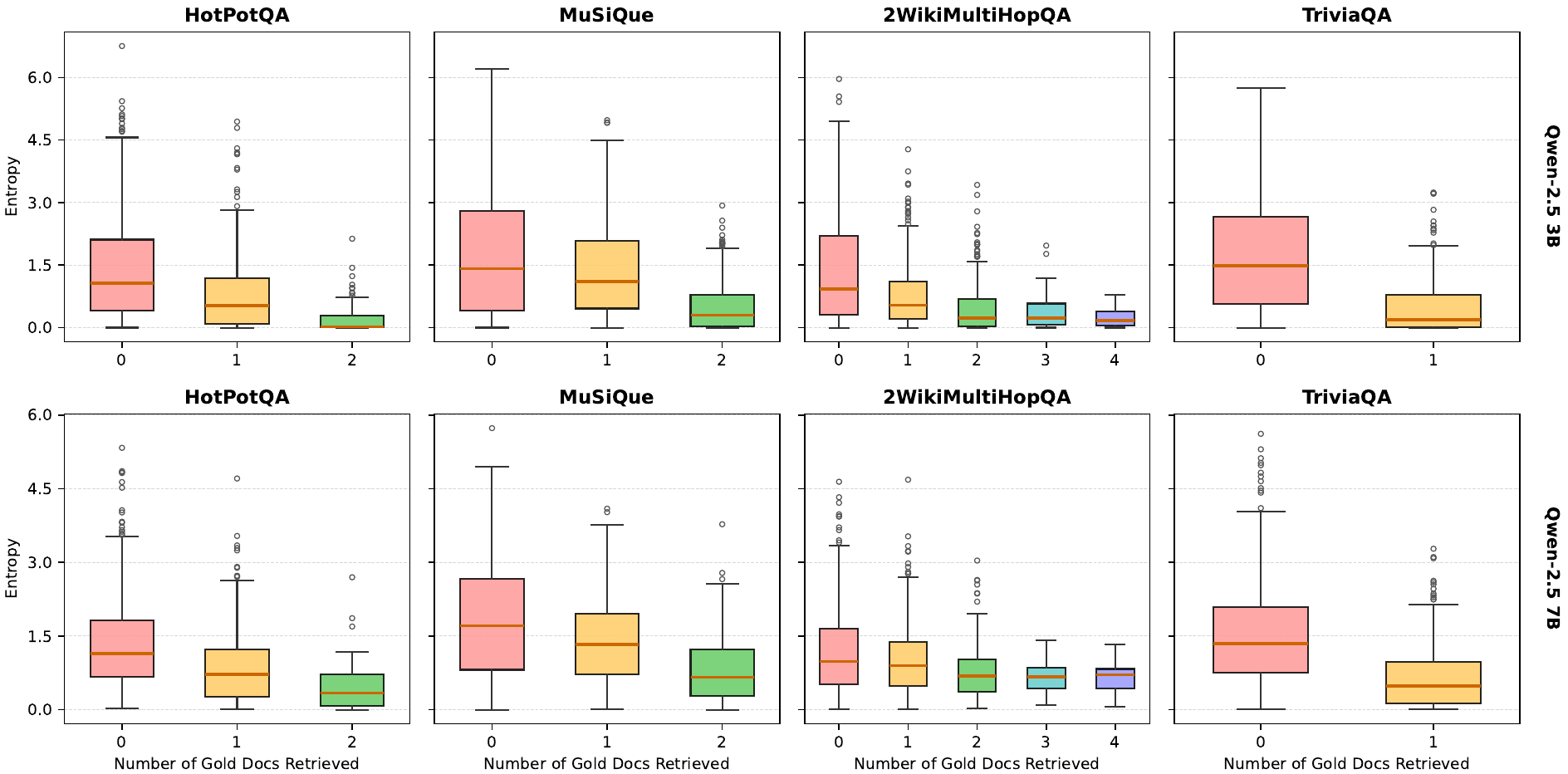}
    \caption{Entropy distribution on RAG benchmarks using Qwen models when answers are generated with gold documents provided in context. Increasing the number of gold documents leads to lower entropy, indicating more confident predictions.}
    \label{fig:entropy_dist}
\end{figure}

Existing iterative RAG systems rely on explicit token-level signals to determine retrieval termination \cite{ircot,searcho1,searchr1}, entangling the stopping decision with surface-level text generation rather than the model's internal assessment of evidence sufficiency. This raises a natural question: \textit{Do the hidden states of an LLM already encode a reliable signal of whether the retrieved context is adequate, even before any answer tokens are produced?}

\paragraph{Setup.} Given an input query token sequence $x$ with an instruction to produce an answer, and a subset $\mathcal{G}_k \subseteq \mathcal{P}$ of $k$ gold supporting documents sampled from the full set $\mathcal{P}$, we construct a probe sequence that enforces answer generation conditioned on the provided context:
\begin{equation}
    \tilde{x}_k = \big(x,\ \mathcal{G}_k,\ \texttt{[PRED]}\big),
\end{equation}
By causal attention, the hidden state $h_k := h_{\texttt{[PRED]}}(\tilde{x}_k) \in \mathbb{R}^d$ attends to the full query and gold context. The LM head induces a distribution over the first answer token, $p_k(v) = \mathrm{softmax}(W h_k)_v$, with predictive entropy: $H_k = -\sum_{v \in \mathcal{V}} p_k(v) \log p_k(v)$, where $\mathcal{V}$ denotes the vocabulary.

\paragraph{Experimental Findings.} We vary $k$ from $0$, as no gold evidence to $k_{\max} = |\mathcal{P}|$, as all gold evidence) and measure the resulting entropy distribution. Figure~\ref{fig:entropy_dist} reports this across four multi-hop QA benchmarks: HotPotQA, TriviaQA, MuSiQue, and 2WikiMultiHopQA, using model Qwen-2.5 Instruct at the 3B and 7B scales without any finetuning. Across all datasets and model combinations, $H_k$ decreases monotonically as $k$ increases. At $k{=}0$, the model exhibits high entropy, reflecting substantial uncertainty over the answer space; at $k{=}k_{\max}$, the distribution concentrates near $H_k < 1.0$. This separation is most pronounced on compositional multi-hop benchmarks, such as MuSiQue and 2WikiMultiHopQA, where the absence of bridge documents prevents shortcuts via parametric recall.

\begin{observation}[Entropy-Sufficiency Correlation]
\label{obs:entropy_sufficiency}
$H_k$ decreases monotonically in $k$; in particular, $\mathbb{E}[H_k \mid \mathcal{P} \subseteq \mathcal{G}_k] < \mathbb{E}[H_k \mid \mathcal{P} \not\subseteq \mathcal{G}_k]$.
\end{observation}

\paragraph{Implication for retrieval control.} Observation~\ref{obs:entropy_sufficiency} establishes that $H_k$ is a reliable signal of evidence sufficiency. Moreover, $H_k$ is a deterministic function of the same hidden state $h_k$ used for retrieval, as in Eq.~\ref{eq:latent_query}, so entropy-based stopping decisions can, in principle, be inferred from $h_k$ alone, without requiring full text generation at each retrieval step. Building on this, Section~\ref{sec:full_method} introduces a lightweight MLP-based Retrieval Control Head that operates directly on $h_{\texttt{[PRED]}}$ to predict whether further retrieval is needed.

\section{Control-Head Supervision under Varying Annotation Regimes}
\label{appen:control_supervision}


The control-head label $y^{(r)}$ defined in Eq.~\ref{eq:control_label} requires access to a set of gold supporting documents $\mathcal{P}$. Such annotations are directly available in standard multi-hop QA datasets (e.g., HotpotQA, 2WikiMultihopQA, MuSiQue), and are the same kind of relevance annotations used to construct positive pairs when training supervised dense retrievers such as DPR, E5, and BGE \citep{karpukhin2020dense, E5, bge-m3}. LAnR uses these annotations both for the contrastive retrieval objective and to derive the stopping label, so the controller introduces no supervision signal beyond what retriever training already requires.

\paragraph{Span-containment labeling for single-hop datasets.}
For single-hop datasets (e.g., NQ, TriviaQA), passage-level annotations are not directly aligned with the retrieval corpus. We adopt a standard Dense Passage Retrieval (DPR)~\citep{karpukhin2020dense} strategy to construct $\mathcal{P}$.

\textbf{Chunking.} We first segment the Wikipedia corpus $\mathcal{C}$ into fixed-length passages (100-word chunks), which serve as the retrieval units. This preprocessing step is deterministic and independent of model training.

\textbf{Positive passage assignment.} We then construct a single positive passage per query using dataset-specific heuristics:

\emph{(i) Datasets with annotated contexts (e.g., NQ).}
We align the original annotated gold passage to the closest matching chunk in $\mathcal{C}$ based on answer span overlap. The matched chunk is treated as the positive passage. Examples for which no alignment is found are discarded.

\emph{(ii) Datasets without annotated contexts (e.g., TriviaQA).}
We adopt distant supervision by retrieving passages using a sparse retriever (e.g., BM25, E5), and selecting the highest-ranked passage that contains the answer string as the positive passage. If no retrieved passage within the top-$k$ results contains the answer, the example is removed.

This procedure yields weakly supervised "gold" passages that are not exhaustively annotated but have been shown to be effective in prior work. The resulting set $\mathcal{P}$ is used consistently for both retrieval training and control-head supervision, ensuring a unified notion of relevance without requiring additional manual annotation.

\section{More Experimental Results}\label{appen:more_results}

\subsection{Retrieval Control Head Analysis}\label{appen:retrieval_control_head_analysis}
\begin{figure}[ht]
    \centering
    \begin{subfigure}[b]{0.48\textwidth}
        \centering
        \includegraphics[width=\textwidth]{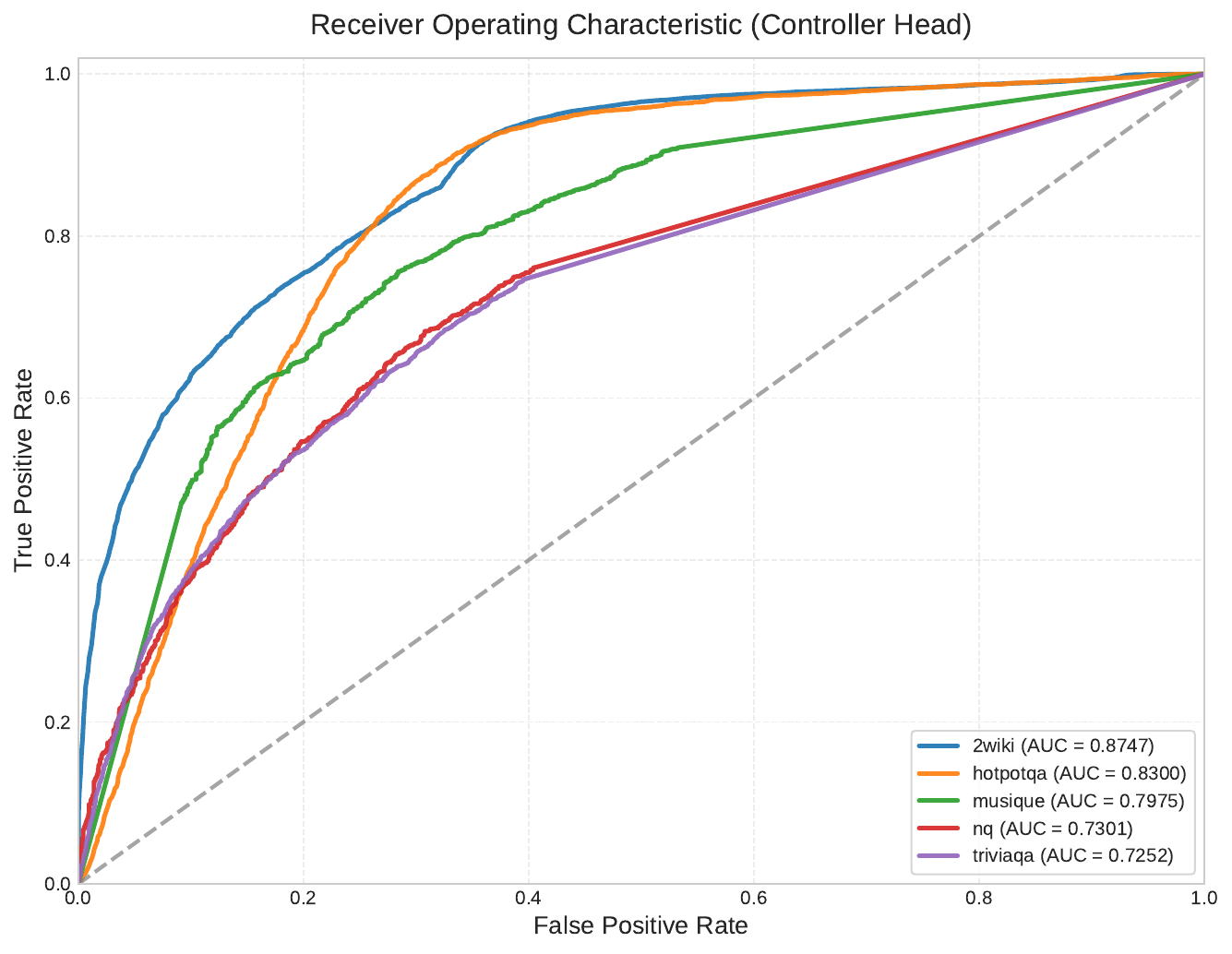}
        \caption{LAnR-3B-Base}
        \label{fig:roc_ctrl_base}
    \end{subfigure}
    \hfill
    \begin{subfigure}[b]{0.48\textwidth}
        \centering
        \includegraphics[width=\textwidth]{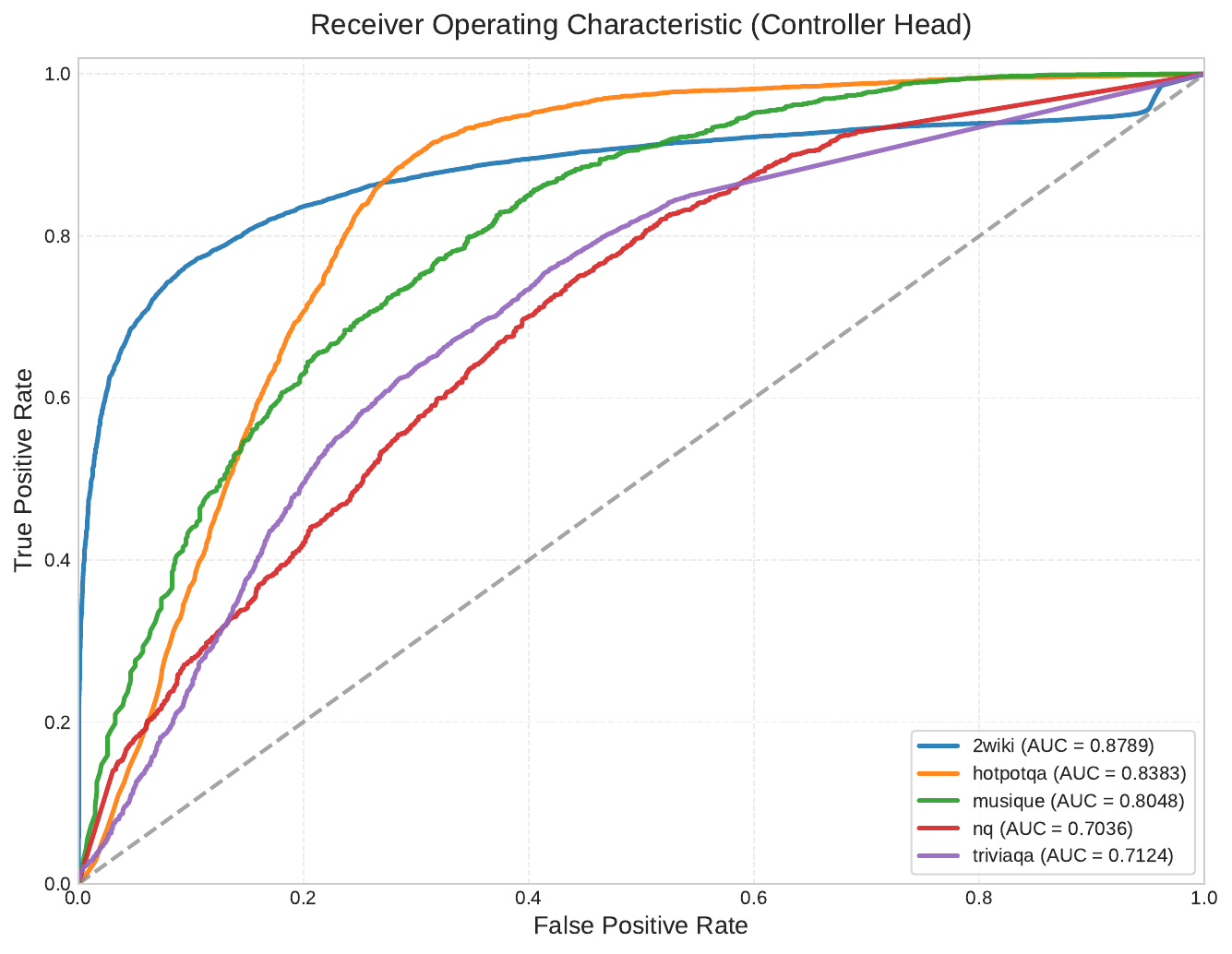}
        \caption{LAnR-3B-Instruct}
        \label{fig:roc_ctrl_instruct}
    \end{subfigure}
    \caption{ROC curves for the retrieval control head across five benchmarks. Instruction tuning yields consistent gains, with the largest improvement on MuSiQue}
    \label{fig:roc_ctrl}
\end{figure}

Figure~\ref{fig:roc_ctrl} shows ROC curves for the retrieval control head on both LAnR-3B-Base and LAnR-3B-Instruct across all five benchmarks.
Both variants achieve strong AUC on multi-hop datasets, confirming that the \texttt{[PRED]} hidden state carries a reliable evidence-sufficiency signal when supporting documents are compositionally distributed.
Performance of both model variants is lower on single-hop datasets (Base: 0.730, 0.725; Instruct: 0.704, 0.712), where the distinction between “enough” and “not enough” evidence is less clear because a single retrieved document often suffices.
The consistent improvement of the Instruct variant across all five datasets and the clear stratification by task complexity confirm that the control head learns a genuine sufficiency signal rather than a dataset-specific heuristic.









\subsection{Effect of Model Size}\label{appen:model_size}
\begin{table}[ht]
\centering
\small
\caption{EM and F1 across methods and model sizes (Qwen-2.5-Base, 3B and 7B) on five QA benchmarks.}
\label{tab:model_size_ablation}
\begin{tabular}{llcccccc}
\toprule
& & \multicolumn{2}{c}{Single-Hop} & \multicolumn{3}{c}{Multi-Hop} & \\
\cmidrule(lr){3-4} \cmidrule(lr){5-7}
Model & Metric & NQ & TriviaQA & HotpotQA & 2Wiki & MuSiQue & Avg. \\
\midrule
\rowcolor{gray!15}
\multicolumn{8}{l}{\emph{Qwen-2.5-3B-Base}} \\
\addlinespace[2pt]
\multirow{2}{*}{SearchR1-3B-Base}
 & EM & 0.421 & 0.583 & 0.297 & 0.274 & 0.066 & 0.328 \\
 & F1 & 0.476 & 0.650 & 0.380 & 0.322 & 0.123 & 0.390 \\
\midrule
\multirow{2}{*}{AutoRefine-3B-Base}
 & EM & 0.467 & 0.620 & 0.405 & 0.393 & 0.157 & 0.408 \\
 & F1 & 0.534 & 0.689 & 0.503 & 0.453 & 0.233 & 0.482 \\
\midrule
\multirow{2}{*}{LAnR-3B-Base (ours)}
 & EM & 0.455 & 0.610 & 0.417 & 0.402 & 0.183 & 0.410 \\
 & F1 & 0.535 & 0.654 & 0.589 & 0.603 & 0.256 & 0.527 \\
\midrule
\rowcolor{gray!15}
\multicolumn{8}{l}{\emph{Qwen-2.5-7B-Base}} \\
\addlinespace[2pt]
\multirow{2}{*}{SearchR1-7B-Base}
 & EM & 0.469 & 0.627 & 0.410 & 0.272 & 0.173 & 0.390 \\
 & F1 & 0.552 & 0.700 & 0.517 & 0.327 & 0.236 & 0.466 \\
\midrule
\multirow{2}{*}{AutoRefine-7B-Base}
 & EM & 0.484 & 0.659 & 0.451 & 0.405 & 0.187 & 0.437 \\
 & F1 & 0.574 & 0.729 & 0.573 & 0.607 & 0.283 & 0.553 \\
\midrule
\multirow{2}{*}{LAnR-7B-Base (ours)}
 & EM & 0.482 & 0.631 & 0.459 & 0.415 & 0.205 & 0.438 \\
 & F1 & 0.570 & 0.724 & 0.590 & 0.612 & 0.295 & 0.558 \\
\bottomrule
\end{tabular}%
\end{table}

\begin{table}[ht]
\scriptsize
\centering
\caption{Statistical analysis against search-during-think baselines. The \textit{p}-value column represents the T-test result of LAnR-Instruct v.s. AutoRefine-Instruct and Search-R1-Instruct.}
\label{tab:statistical_analysis}
\setlength{\tabcolsep}{3pt}
\begin{tabular}{lccccc c}
\hline
Model & NQ & TriviaQA & HotpotQA & 2wiki & Musique & $p$-value \\
\hline
LAnR-Instruct
& $0.460 \pm 0.008$
& $0.613 \pm 0.005$
& $0.419 \pm 0.006$
& $0.408 \pm 0.006$
& $0.193 \pm 0.005$
& $-$ \\

AutoRefine-Instruct
& $0.461 \pm 0.010$
& $0.620 \pm 0.007$
& $0.405 \pm 0.012$
& $0.404 \pm 0.010$
& $0.145 \pm 0.011$
& $7.49 \times 10^{-3}$  \\

Search-R1-Instruct
& $0.410 \pm 0.009$
& $0.597 \pm 0.019$
& $0.315 \pm 0.016$
& $0.254 \pm 0.023$
& $0.124 \pm 0.005$
& $1.31 \times 10^{-39}$  \\
\hline
\end{tabular}
\end{table}

\begin{figure}[ht]
    \centering
    \includegraphics[width=\textwidth]{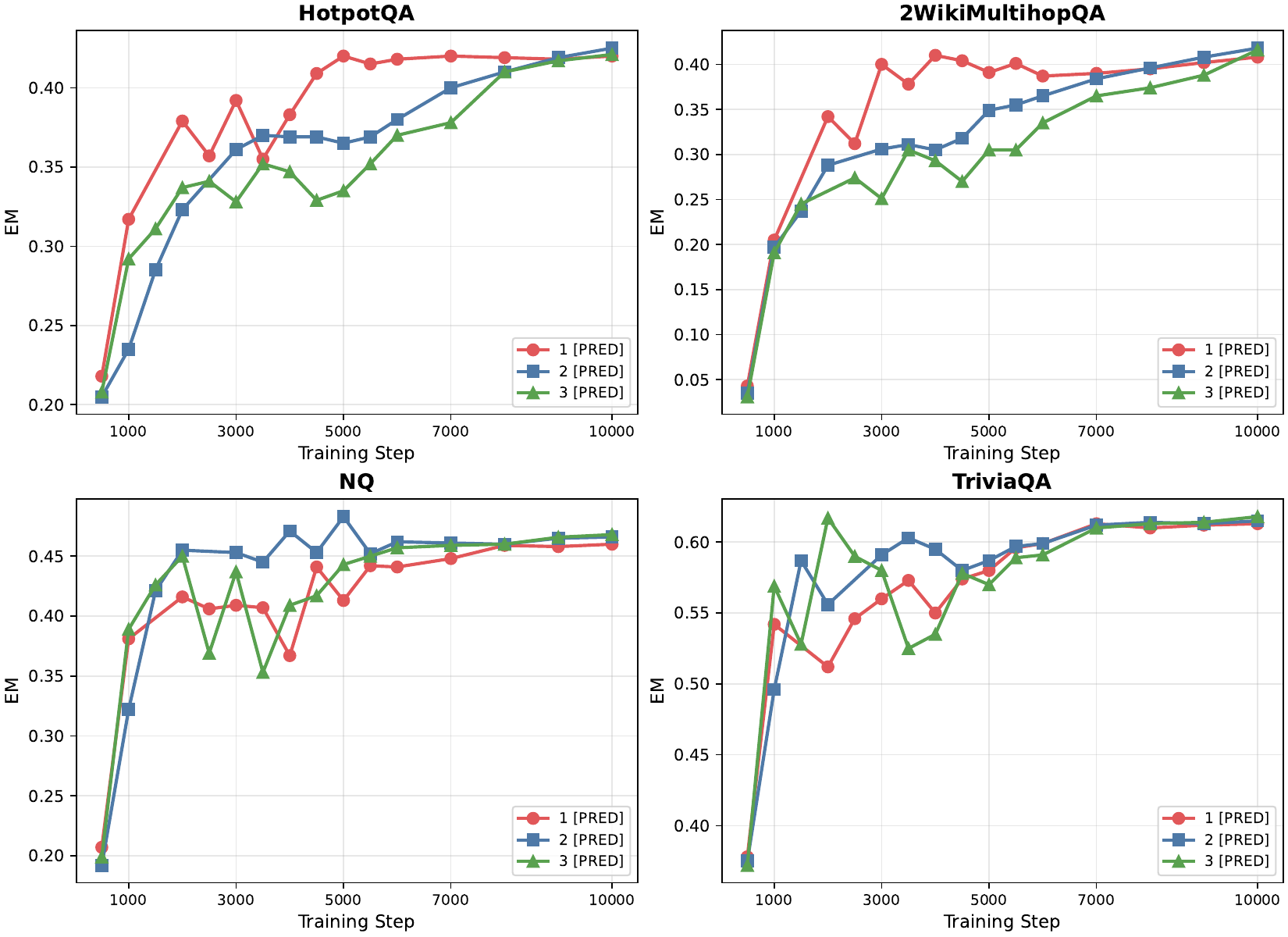}
    \caption{EM accuracy by training step for models trained with 1, 2, and 3 \texttt{[PRED]} tokens across four benchmarks. $N{=}1$ converges fastest and performs best on multi-hop datasets; additional tokens offer marginal gains only on single-hop TriviaQA.}
    \label{fig:pred_ablation}
\end{figure}

Table~\ref{tab:model_size_ablation} compares Search-R1, AutoRefine, and LAnR across Qwen-2.5-Base 3B and 7B backbones using both EM and F1 metrics. Scaling from 3B to 7B consistently improves performance for all methods, with the largest gains observed on multi-hop benchmarks that require compositional reasoning. At both model scales, LAnR achieves the strongest overall multi-hop performance. With the 3B backbone, LAnR-3B-Base attains the highest EM on HotpotQA (0.417), 2Wiki (0.402), and MuSiQue (0.183), while also substantially outperforming prior methods in F1, particularly on HotpotQA (0.589 vs.\ 0.503 for AutoRefine) and 2Wiki (0.603 vs.\ 0.453). Similar trends hold at 7B scale, where LAnR-7B-Base achieves the best EM and F1 across all three multi-hop datasets, reaching 0.459/0.590 on HotpotQA, 0.415/0.612 on 2Wiki, and 0.205/0.295 on MuSiQue. Although AutoRefine remains competitive on single-hop datasets such as NQ and TriviaQA, the results show that LAnR benefits more consistently from increased model capacity, suggesting that latent retrieval scales effectively with stronger backbone representations and is particularly advantageous for complex multi-step reasoning tasks.


\subsection{Variance Analysis}\label{appen:variance_analysis}
Table~\ref{tab:statistical_analysis} reports the mean and standard deviation across three runs with different random seeds to evaluate the robustness of search-during-think methods. LAnR-Instruct consistently achieves lower variance than competing baselines across most datasets, indicating more stable retrieval and reasoning behavior. To assess statistical significance, we perform paired T-tests between LAnR-Instruct and AutoRefine-Instruct. The resulting low $p$-values ($p \ll 0.01$) confirm that the performance differences are statistically significant, demonstrating that the gains of latent retrieval are reliable rather than arising from random variation.

\subsection{Ablation on Numbers of \texttt{[PRED]} Tokens}\label{appen:PRED_numers}

We further analyze the effect of training with multiple \texttt{[PRED]} tokens, detailed in Figure \ref{fig:pred_ablation}. Increasing the number of \texttt{[PRED]} tokens encourages the model to construct more abstract latent queries. On single-hop benchmarks such as TriviaQA and NQ, LAnR consistently benefits from additional \texttt{[PRED]} tokens, with performance improvements emerging early in training. In contrast, on more challenging multi-hop benchmarks, models with multiple \texttt{[PRED]} tokens require substantially longer training before surpassing the performance of the single-\texttt{[PRED]} setting. These results suggest that while additional \texttt{[PRED]} tokens can improve abstraction capability, they also increase optimization difficulty for complex reasoning tasks. Therefore, we adopt $N=1$ as the default configuration throughout our experiments.

\subsection{Comparison with Adaptive RAG Methods}\label{appen:adaptive_rag}
 
LAnR decides when to continue or stop retrieving, so we additionally compare it with two representative adaptive RAG methods that also learn retrieval timing: Self-RAG \cite{self-rag}, and CtrlA \cite{huanshuo2025ctrla}. Both baselines use the same E5 retriever over the 2018 Wikipedia corpus as our other baselines.
 
\paragraph{Metrics.} In addition to Exact Match (EM), we report Contains-Match (CM), which considers a prediction correct if it contains the normalized gold answer as a span. Since Self-RAG and CtrlA often generate verbose, free-form responses instead of concise extractive answers, EM may substantially underestimate their performance. For instance, Self-RAG achieves an EM of 0.215 but a CM of 0.730 on TriviaQA. CM provides a more permissive evaluation criterion that partially mitigates this discrepancy in answer formats.
 
\begin{table}[ht]
\centering
\small
\setlength{\tabcolsep}{4pt}
\caption{\textbf{Comparison with adaptive RAG methods.} EM and Contains-Match (CM) on five QA benchmarks. Self-RAG and CtrlA use 7B backbones (Llama-2-7B and Mistral-7B-Instruct); LAnR uses Qwen2.5-3B and Qwen2.5-7B backbones. All methods retrieve from the same 2018 Wikipedia corpus.}
\label{tab:adaptive_rag}
\begin{tabular}{@{}llcccccc@{}}
\toprule
& & \multicolumn{2}{c}{Single-Hop} & \multicolumn{3}{c}{Multi-Hop} & \\
\cmidrule(lr){3-4} \cmidrule(lr){5-7}
Model & Metric & NQ & TriviaQA & HotpotQA & 2Wiki & MuSiQue & Avg. \\
\midrule
\multirow{2}{*}{Self-RAG (7B)}
 & EM & 0.230 & 0.215 & 0.145 & 0.015 & 0.025 & 0.126 \\
 & CM & 0.455 & \textbf{0.730} & 0.385 & 0.350 & 0.090 & 0.402 \\
\midrule
\multirow{2}{*}{CtrlA (7B)}
 & EM & 0.125 & 0.120 & 0.113 & 0.110 & 0.131 & 0.120 \\
 & CM & 0.535 & 0.710 & 0.370 & 0.400 & 0.160 & 0.435 \\
\midrule
\rowcolor{gray!15}
 & EM & 0.455 & 0.610 & 0.417 & 0.402 & 0.187 & 0.414 \\
\rowcolor{gray!15}
\multirow{-2}{*}{LAnR-Base-3B}
 & CM & 0.540 & 0.693 & 0.502 & 0.472 & 0.248 & 0.491 \\
\midrule
\rowcolor{gray!15}
 & EM & \textbf{0.482} & \textbf{0.631} & \textbf{0.459} & \textbf{0.415} & \textbf{0.205} & \textbf{0.438} \\
\rowcolor{gray!15}
\multirow{-2}{*}{LAnR-Base-7B}
 & CM & \textbf{0.597} & 0.719 & \textbf{0.514} & \textbf{0.494} & \textbf{0.266} & \textbf{0.518} \\
\bottomrule
\end{tabular}
\end{table}
 
\paragraph{Results.} Table~\ref{tab:adaptive_rag} shows that both LAnR variants outperform Self-RAG and CtrlA on all benchmarks under EM, including LAnR-Base-3B, which uses fewer than half as many parameters. Under CM, LAnR-Base-7B achieves the best average score (0.518) and leads on four of five benchmarks. Self-RAG performs best on TriviaQA (0.730 vs.\ 0.719), where LAnR-Base-3B also slightly trails both baselines. LAnR's gains are particularly pronounced on multi-hop tasks, as shown on MuSiQue (0.248 vs.\ 0.090 for Self-RAG and 0.160 for CtrlA). These results suggest that the improvements stem not merely from model scale, but from jointly optimizing latent-space retrieval control and query formation with answer generation. 
 
\paragraph{Retrieval recall against iterative and adaptive methods.} To compare retrieval quality directly, we evaluate all methods under a controlled budget of two sequential retrieval rounds with three documents per round (six documents in total) and report Recall@6 on the three multi-hop benchmarks. All baselines use the same E5 retriever and 2018 Wikipedia index, so they differ only in how they form retrieval queries and decide when to retrieve; LAnR retrieves over the same corpus with its own latent queries and document embeddings. We also include \emph{Naive 2-hop E5}, a retrieval-only baseline without LLM involvement: after retrieving the top-3 passages for the question, snippets from these passages are appended to the question to form the second-round query.
 
\begin{table}[ht]
\centering
\small
\caption{\textbf{Retrieval recall (Recall@6) of adaptive and iterative methods} under a budget of two retrieval rounds $\times$ top-3 documents. All baselines use the same E5 retriever and Wikipedia index.}
\label{tab:adaptive_recall}
\begin{tabular}{@{}lccc@{}}
\toprule
Method & HotpotQA & 2Wiki & MuSiQue \\
\midrule
Naive 2-hop E5 (reformulation) & 0.805 & 0.662 & 0.446 \\
Search-R1  \cite{searchr1}     & 0.828 & 0.674 & 0.468 \\
CtrlA      \cite{huanshuo2025ctrla} & 0.832 & 0.680 & 0.496 \\
Self-RAG   \cite{self-rag}     & 0.842 & 0.717 & 0.492 \\
AutoRefine \cite{auto-refine}  & 0.853 & 0.706 & 0.503 \\
\midrule
\rowcolor{gray!15}
LAnR & \textbf{0.865} & \textbf{0.724} & \textbf{0.520} \\
\bottomrule
\end{tabular}
\end{table}
 
Table~\ref{tab:adaptive_recall} shows that LAnR achieves the highest recall on all three benchmarks under the same budget. Whereas text-based methods must first generate an intermediate query, which can introduce ambiguity or lexical bias, each LAnR query is optimized directly toward the remaining evidence set $\mathcal{P}^{(r)}=\mathcal{P}\setminus\mathcal{R}^{(r-1)}$ through the multi-turn contrastive objective (Eq.~\ref{eq:adaptive_cl}).

\subsection{Query and Document Read-out Design}\label{appen:readout_design}
 
LAnR reads out the query from a dedicated \texttt{[PRED]} token and the document embedding from the last token of the document (Eqs.~\ref{eq:latent_query}--\ref{eq:doc_repr}). The goal of this design is to let one model produce both representations while limiting interference between the retrieval and generation objectives. The asymmetry lies only in the read-out position: both embeddings come from the same LLM parameters and are optimized jointly with the same contrastive objective.

\begin{table}[ht]
\centering
\small
\caption{EM over training steps when the last input token (no \texttt{[PRED]} token) is used as the retrieval query.}
\label{tab:no_pred}
\begin{tabular}{@{}rcccc@{}}
\toprule
Steps & NQ & TriviaQA & HotpotQA & 2Wiki \\
\midrule
1000 & 0.103 & 0.247 & 0.186 & 0.145 \\
3000 & 0.194 & 0.298 & 0.224 & 0.182 \\
5000 & 0.236 & 0.314 & 0.259 & 0.213 \\
\bottomrule
\end{tabular}
\end{table}
 
\paragraph{Query side: dedicated \texttt{[PRED]} token vs.\ last input token.} Our initial design used the hidden state of the last input token directly as the query, i.e., zero \texttt{[PRED]} tokens. Table~\ref{tab:no_pred} reports the EM of this variant over training. It trains substantially more slowly and stays well below the \texttt{[PRED]}-based model (Figure~\ref{fig:pred_ablation}). We hypothesize that the last-token state is optimized simultaneously for next-token prediction and retrieval, which creates competing objectives, whereas a dedicated \texttt{[PRED]} position provides a separate read-out location where the model can encode retrieval intent without disturbing its generative representation.
  

\section{Limitations}\label{appen:limitations}

\textbf{Corpus encoding overhead.} LAnR encodes the entire retrieval corpus using the same LLM used for generation. While this enables tight representational alignment, it is more computationally expensive than conventional retrieval approaches,
including sparse lexical methods such as BM25
\cite{robertson2009probabilistic} and lightweight dense retrievers such as
BGE \cite{bge-m3}. Re-encoding is also required whenever the backbone model is updated, which may limit practicality in frequently updated knowledge bases.

\textbf{Interpretability of latent queries.} Unlike text-based RAG systems, where retrieval queries are human-readable and easily auditable, LAnR's latent query vectors are not directly interpretable. This makes it harder to diagnose retrieval failures or explain why specific documents were retrieved, which may be a concern in high-stakes deployments.

\textbf{Evaluation scope.} All experiments use English-language open-domain QA benchmarks with Wikipedia as the knowledge source. Performance on non-English, domain-specific, or long-document corpora remains untested, and results may not generalize to settings with significantly different document distributions or answer styles.

\section{Broader Impact}\label{appen:broader_impact}
LAnR is foundational research on latent retrieval-augmented generation efficiency and does not target any specific deployment. Nevertheless, we briefly discuss potential societal implications.

\textbf{Positive impacts.} By reducing inference latency by up to $2.7\times$ and token generation by $\approx30\times$ compared to existing RAG systems, LAnR lowers the computational cost of knowledge-grounded language model inference, improving accessibility for resource-constrained practitioners and reducing energy consumption at scale. Improved factuality in question answering also has broad benefits in educational, medical, and scientific information access.

\textbf{Potential negative impacts.} As with any system that improves the fluency and factual grounding of LLM outputs, LAnR could lower the barrier to generating more convincing disinformation or fabricated content at scale. The efficiency gains in particular may make high-volume automated generation more feasible. Additionally, because LAnR retrieves from large external corpora, any biases present in those corpora may be silently propagated into generated answers without explicit query traces, potentially making such biases harder to audit than in text-based retrieval pipelines where queries are inspectable.

\section{Case Studies}\label{appen:case_study}

The following examples trace LAnR's retrieval process hop by hop. At each search, the model runs one forward pass over a growing context ending with \texttt{[PRED]}. The hidden state at that position encodes all evidence seen so far, drives the next dense retrieval query, and is fed to the MLP control head to decide whether to continue. The context after $h$ searches is:
\[
\texttt{[Q]}\;\textit{question}\;
\underbrace{[d_1^{(1)}]\cdots[d_K^{(1)}]}_{\text{search 1}}\;\cdots\;
\underbrace{[d_1^{(r)}]\cdots[d_K^{(r)}]}_{\text{search }r}\;
\texttt{[PRED]}
\]
Gold supporting documents are \underline{underlined} with~$^\star$.

\clearpage
\newpage
\subsection*{Example 1 - MuSiQue (3-hop)}

\begin{tcolorbox}[
    colback=gray!4, colframe=gray!35, arc=2pt,
    title={\textbf{2WikiMultihopQA} \textnormal{-} \textbf{2-hop compositional}},
    fonttitle=\small\bfseries, left=5pt, right=5pt, top=5pt, bottom=5pt]

\textbf{Question:} \textit{"Where do Greyhound buses leave from in the city where the performer of Darlings formed?"}\\[2pt]
\textbf{Reasoning chain:} Darlings $\to$ performer: Kevin Drew $\to$ formation city: Toronto $\to$ bus terminal

\vspace{0.5em}
\noindent\textbf{Turn 1}\hfill$p_{\mathrm{ctrl}}=0.9977\;\to\;\textsc{continue}$

\smallskip\noindent
\textbf{Context:}\enspace{\small%
\textcolor{blue!60!black}{\texttt{[Q]}}\;%
\textit{Where do Greyhound buses leave from\,$\ldots$\,formed?}\;%
\textcolor{blue!60!black}{\textbf{\texttt{[PRED]}}}}

\smallskip\noindent\textbf{Retrieved:} \underline{Kevin Drew}$^{\star}$,\; Spirit If$\ldots$,\; Detropia

\smallskip\noindent\textcolor{gray!65!black}{\footnotesize%
\textbf{[Kevin Drew]}\enspace
Kevin Drew (born September 9, 1976) is a Canadian musician and songwriter who, together with
Brendan Canning, founded the expansive Toronto baroque-pop collective Broken Social Scene.}

\vspace{0.55em}
\noindent\textbf{Turn 2}\hfill$p_{\mathrm{ctrl}}=0.7922\;\to\;\textsc{continue}$

\smallskip\noindent
\textbf{Context:}\enspace{\small%
\textcolor{blue!60!black}{\texttt{[Q]}}\;%
\textit{$\ldots$}\;%
\textcolor{teal!60!black}{\texttt{[Kevin Drew: \textit{$\ldots$Toronto baroque-pop collective$\ldots$}]}}\;%
\texttt{[Spirit If:\,$\cdots$]\;[Detropia:\,$\cdots$]}\;%
\textcolor{blue!60!black}{\textbf{\texttt{[PRED]}}}}

\smallskip\noindent\textbf{Retrieved:} \underline{Toronto Coach Terminal}$^{\star}$,\; Alvarado Transportation Center,\; Dufferin St.\ Bridges

\smallskip\noindent\textcolor{gray!65!black}{\footnotesize%
\textbf{[Toronto Coach Terminal]}\enspace
The Toronto Coach Terminal is the central bus station for inter-city services in Toronto, Ontario,
Canada. It is located at 610 Bay Street, in the city's Downtown.}

\vspace{0.55em}
\noindent\textbf{Turn 3}\hfill$p_{\mathrm{ctrl}}\approx 0\;\to\;\textsc{stop}$

\smallskip\noindent
\textbf{Context:}\enspace{\small%
\textcolor{blue!60!black}{\texttt{[Q]}}\;%
\textit{$\ldots$}\;%
\textcolor{teal!60!black}{\texttt{[Kevin Drew:\,$\cdots$]\;[Toronto Coach Terminal: \textit{$\ldots$Toronto$\ldots$Bay St$\ldots$}]}}\;%
\texttt{$\cdots$}\;%
\textcolor{blue!60!black}{\textbf{\texttt{[PRED]}}}}

\smallskip\noindent\textbf{Retrieved:} \underline{Darlings (Kevin Drew album)}$^{\star}$,\; Peace Center,\; Darlington County

\smallskip\noindent\textcolor{gray!65!black}{\footnotesize%
\textbf{[Darlings (Kevin Drew album)]}\enspace
Darlings is the second solo album by Broken Social Scene co-founder Kevin Drew.
It was released on March 18, 2014.}

\vspace{0.6em}
\noindent\textbf{Answer:}\;\texttt{The Toronto Coach Terminal}\quad\textcolor{green!45!black}{$\checkmark$\;\textit{Correct}}\hfill\footnotesize$^\star$ gold document
\end{tcolorbox}

The \texttt{[PRED]} state at Search 1 already encodes \textit{Toronto} implicitly through the Kevin Drew article, letting the model jump directly to the bus-terminal query at Search 2 without materialising "Toronto" as an explicit intermediate result.
All three gold documents are retrieved across three latent searches, and the control head stops as soon as the evidence chain is complete.

\clearpage
\newpage
\subsection*{Example 2 - 2WikiMultihopQA (2-hop)}

\begin{tcolorbox}[
    colback=gray!4, colframe=gray!35, arc=2pt,
    title={\textbf{2WikiMultihopQA} \textnormal{-} \textbf{2-hop compositional}},
    fonttitle=\small\bfseries, left=5pt, right=5pt, top=5pt, bottom=5pt]

\textbf{Question:} \textit{"Where was the director of film \emph{The Green Fog} born?"}

\textbf{Reasoning chain:} The Green Fog $\to$ director: Guy Maddin $\to$ birthplace: Winnipeg, Manitoba

\vspace{0.5em}
\noindent\textbf{Turn 1}\hfill$p_{\mathrm{ctrl}}=0.9999\;\to\;\textsc{continue}$

\smallskip\noindent
\textbf{Context:}\enspace{\small%
\textcolor{blue!60!black}{\texttt{[Q]}}\;%
\textit{Where was the director of film The Green Fog born?}\;%
\textcolor{blue!60!black}{\textbf{\texttt{[PRED]}}}}

\smallskip\noindent\textbf{Retrieved:} \underline{The Green Fog}$^{\star}$,\; Walon Green,\; Adam Green (filmmaker)

\smallskip\noindent\textcolor{gray!65!black}{\footnotesize%
\textbf{[The Green Fog]}\enspace
The Green Fog is an experimental film directed by Guy Maddin, Evan Johnson, and Galen Johnson
that loosely revisits the plot of Alfred Hitchcock's \textit{Vertigo} through found footage.}

\vspace{0.55em}
\noindent\textbf{Turn 2}\hfill$p_{\mathrm{ctrl}}=0.9992\;\to\;\textsc{continue}$

\smallskip\noindent
\textbf{Context:}\enspace{\small%
\textcolor{blue!60!black}{\texttt{[Q]}}\;%
\textit{$\ldots$}\;%
\textcolor{teal!60!black}{\texttt{[The Green Fog: \textit{$\ldots$directed by Guy Maddin$\ldots$}]}}\;%
\texttt{[Walon Green:\,$\cdots$]\;[Adam Green:\,$\cdots$]}\;%
\textcolor{blue!60!black}{\textbf{\texttt{[PRED]}}}}

\smallskip\noindent\textbf{Retrieved:} \underline{Guy Maddin}$^{\star}$,\; The Heart of the World,\; The Forbidden Room

\smallskip\noindent\textcolor{gray!65!black}{\footnotesize%
\textbf{[Guy Maddin]}\enspace
Guy Maddin (born February 28, 1956) is a Canadian screenwriter, director, author, cinematographer,
and film editor of both features and short films, from \textbf{Winnipeg, Manitoba}.}

\vspace{0.55em}
\noindent\textbf{Turn 3}\hfill$p_{\mathrm{ctrl}}=0.0000\;\to\;\textsc{stop}$

\smallskip\noindent
\textbf{Context:}\enspace{\small%
\textcolor{blue!60!black}{\texttt{[Q]}}\;%
\textit{$\ldots$}\;%
\textcolor{teal!60!black}{\texttt{[The Green Fog:\,$\cdots$]\;[Guy Maddin: \textit{$\ldots$Winnipeg, Manitoba$\ldots$}]}}\;%
\texttt{$\cdots$}\;%
\textcolor{blue!60!black}{\textbf{\texttt{[PRED]}}}}

\smallskip\noindent\textcolor{gray!50!black}{\footnotesize\textit{(Control head halts; no new documents retrieved.)}}

\vspace{0.6em}
\noindent\textbf{Answer:}\;\texttt{Winnipeg, Manitoba}\quad\textcolor{green!45!black}{$\checkmark$\;\textit{Correct}}
\hfill\footnotesize$^\star$ gold document
\end{tcolorbox}

This example shows the sharpest control-head transition across all three datasets: $p_{\mathrm{ctrl}}$ holds at $0.9999$ and $0.9992$ while evidence accumulates, then drops to $0.0000$ the moment \textit{Guy Maddin} and his birthplace are both in context.
No text query names "Guy Maddin" at any step; the latent query at Search 2 is shaped entirely by the \texttt{[PRED]} representation formed over the \textit{The Green Fog} article.

\newpage
\section*{NeurIPS Paper Checklist}

\begin{enumerate}

\item {\bf Claims}
    \item[] Question: Do the main claims made in the abstract and introduction accurately reflect the paper's contributions and scope?
    \item[] Answer: \answerYes{} 
    \item[] Justification: The main claims made in the abstract and introduction accurately reflect the main paper's contributions and scope.

\item {\bf Limitations}
    \item[] Question: Does the paper discuss the limitations of the work performed by the authors?
    \item[] Answer: \answerYes{} 
    \item[] Justification: We include a dedicated limitation discussion in Appendix \ref{appen:limitations}.

\item {\bf Theory assumptions and proofs}
    \item[] Question: For each theoretical result, does the paper provide the full set of assumptions and a complete (and correct) proof?
    \item[] Answer: \answerNA{} 
    \item[] Justification: There are no theoretical results in this paper.

    \item {\bf Experimental result reproducibility}
    \item[] Question: Does the paper fully disclose all the information needed to reproduce the main experimental results of the paper to the extent that it affects the main claims and/or conclusions of the paper (regardless of whether the code and data are provided or not)?
    \item[] Answer: \answerYes{} 
    \item[] Justification: The paper fully discloses all the information needed to reproduce the main experimental results of the paper in Appendix \ref{appen:detail_training}, and we provide the code to assist the reproducibility of our experimental results in \url{https://github.com/akBear23/LAnR}.

\item {\bf Open access to data and code}
    \item[] Question: Does the paper provide open access to the data and code, with sufficient instructions to faithfully reproduce the main experimental results, as described in supplemental material?
    \item[] Answer: \answerYes{} 
    \item[] Justification: This paper provides open access to the code, includes instructions for running the code in the link \url{https://github.com/akBear23/LAnR}, and we use public data to train in link \url{https://huggingface.co/datasets/RUC-NLPIR/FlashRAG_datasets}.

\item {\bf Experimental setting/details}
    \item[] Question: Does the paper specify all the training and test details (e.g., data splits, hyperparameters, how they were chosen, type of optimizer) necessary to understand the results?
    \item[] Answer: \answerYes{} 
    \item[] Justification: Appendix \ref{appen:detail_training} contains all experimental details.
    \item[] Guidelines:
    \begin{itemize}
        \item The answer \answerNA{} means that the paper does not include experiments.
        \item The experimental setting should be presented in the core of the paper to a level of detail that is necessary to appreciate the results and make sense of them.
        \item The full details can be provided either with the code, in the appendix, or as supplemental material.
    \end{itemize}

\item {\bf Experiment statistical significance}
    \item[] Question: Does the paper report error bars suitably and correctly defined or other appropriate information about the statistical significance of the experiments?
    \item[] Answer: \answerYes{} 
    \item[] Justification: We provide statistical significance analyses of different seeds in Appendix \ref{appen:variance_analysis}.

\item {\bf Experiments compute resources}
    \item[] Question: For each experiment, does the paper provide sufficient information on the computer resources (type of compute workers, memory, time of execution) needed to reproduce the experiments?
    \item[] Answer: \answerYes{} 
    \item[] Justification:  We include all experiments computed in the Appendix \ref{appen:detail_training}, covering all resources used for generating model responses.
    
\item {\bf Code of ethics}
    \item[] Question: Does the research conducted in the paper conform, in every respect, with the NeurIPS Code of Ethics \url{https://neurips.cc/public/EthicsGuidelines}?
    \item[] Answer: \answerYes{} 
    \item[] Justification: We confirm the research conducted in the paper conforms, in every respect, with the NeurIPS Code of Ethics.

\item {\bf Broader impacts}
    \item[] Question: Does the paper discuss both potential positive societal impacts and negative societal impacts of the work performed?
    \item[] Answer: \answerYes{} 
    \item[] Justification: We discuss broader impact in Appendix \ref{appen:broader_impact}.
    
\item {\bf Safeguards}
    \item[] Question: Does the paper describe safeguards that have been put in place for responsible release of data or models that have a high risk for misuse (e.g., pre-trained language models, image generators, or scraped datasets)?
    \item[] Answer: \answerNA{} 
    \item[] Justification: The paper poses no such risks.

\item {\bf Licenses for existing assets}
    \item[] Question: Are the creators or original owners of assets (e.g., code, data, models), used in the paper, properly credited and are the license and terms of use explicitly mentioned and properly respected?
    \item[] Answer: \answerYes{} 
    \item[] Justification: The datasets are used via \url{https://huggingface.co/datasets/RUC-NLPIR/FlashRAG_datasets} under \texttt{CC-BY-SA-4.0} license. We use the Qwen-2.5 series models \cite{yang2025qwen3} that are under the Apache license 2.0.

\item {\bf New assets}
    \item[] Question: Are new assets introduced in the paper well documented and is the documentation provided alongside the assets?
    \item[] Answer: \answerNA{} 
    \item[] Justification: This paper does not release new assets.

\item {\bf Crowdsourcing and research with human subjects}
    \item[] Question: For crowdsourcing experiments and research with human subjects, does the paper include the full text of instructions given to participants and screenshots, if applicable, as well as details about compensation (if any)? 
    \item[] Answer: \answerNA{} 
    \item[] Justification: The paper does not involve crowdsourcing nor research with human subjects.

\item {\bf Institutional review board (IRB) approvals or equivalent for research with human subjects}
    \item[] Question: Does the paper describe potential risks incurred by study participants, whether such risks were disclosed to the subjects, and whether Institutional Review Board (IRB) approvals (or an equivalent approval/review based on the requirements of your country or institution) were obtained?
    \item[] Answer: \answerNA{} 
    \item[] Justification: The paper does not involve crowdsourcing nor research with human subjects.

\item {\bf Declaration of LLM usage}
    \item[] Question: Does the paper describe the usage of LLMs if it is an important, original, or non-standard component of the core methods in this research? Note that if the LLM is used only for writing, editing, or formatting purposes and does \emph{not} impact the core methodology, scientific rigor, or originality of the research, declaration is not required.
    \item[] Answer: \answerNA{} 
    \item[] Justification: The core method development in this research does not involve LLMs as any
important, original, or non-standard components.

\end{enumerate}

\end{document}